\pdfoutput=1
\documentclass[11pt]{article}


\usepackage[preprint]{acl}
\usepackage{algorithm}
\usepackage{times}
\usepackage{latexsym}

\usepackage{inconsolata}
\usepackage{amsmath}

\usepackage{caption}
\usepackage{subcaption}
\usepackage[utf8]{inputenc} % allow utf-8 input
\usepackage[T1]{fontenc}    % use 8-bit T1 fonts
\usepackage{url}            % simple URL typesetting
\usepackage{amsfonts}       % blackboard math symbols
\usepackage{nicefrac}       % compact symbols for 1/2, etc.
\usepackage{microtype}      % microtypography
\usepackage{xcolor}         % colors
\usepackage{wrapfig}
\usepackage{booktabs}
\usepackage{enumitem}
\usepackage{makecell}
\usepackage{diagbox}
\usepackage{bm}
\usepackage{dsfont}
\usepackage{multirow}
\usepackage{varwidth}
\usepackage{pifont}
\usepackage{makecell}
\usepackage{graphicx} 
\usepackage{comment}
\usepackage[colaction]{multicol}
\usepackage{colortbl}
\usepackage{algpseudocode} %%%Zhao: I added this, don't delete it.
\usepackage{xspace}
\usepackage{rotating}
\usepackage{longtable}
\usepackage{adjustbox}
\usepackage{threeparttable}
\usepackage{listings}

\usepackage{hyperref}
\usepackage{bbm}
\usepackage{tabularx}

\definecolor{customblue}{HTML}{2F5C8A}
\definecolor{customred}{HTML}{B5483D}
\definecolor{customgray}{HTML}{85888d}

\newcommand{\aname}{ConSA\xspace}
\title{\aname: Controllable Sparsity in Hybrid Attention via
Learnable Allocation}

\author{
Yao Chen$^{1,2}$\thanks{~denotes equal contribution. $^\dagger$ denotes the corresponding author.},
~Yinqi Yang$^{3*}$,
~Junyuan Shang$^{3}$,
~Xiangzhao Hao$^{3}$,
~Simeng Zhang$^{1,2}$,\\
~\textbf{Yilong Chen}$^{1,2}$,
~\textbf{Tingwen Liu}$^{1,2\dagger}$\textbf{,} 
~\textbf{Shuohuan Wang}$^3$\textbf{,}
~\textbf{Dianhai Yu}$^3$\\
\small  \normalsize $^1$ Institute of Information Engineering, Chinese Academy of Sciences\\
\small  \normalsize $^2$ School of Cyber Security, University of Chinese Academy of Sciences\\
\small  \normalsize $^3$ Baidu Inc.\\
\small \{\texttt{chenyao2023, liutingwen\}@iie.ac.cn} \\
\small \{\texttt{yangyinqi, shangjunyuan, wangshuohuan\}@baidu.com}\\
}

\begin{document}
\maketitle

\begin{abstract}\label{sec.abs}

Hybrid architectures combining full attention (FA) and sliding-window attention (SWA) are a promising paradigm for efficient LLM inference.
However, existing methods typically rely on hand-crafted rules or simple post-hoc heuristics for FA/SWA allocation and offer limited analysis of the attention behaviors underlying these designs.
We propose Controllable Sparsity in Hybrid Attention (ConSA), a framework that learns optimal FA/SWA assignment under a user-specified sparsity target.
ConSA employs L0 regularization to learn binary masks selecting between FA and SWA for each attention unit, while an augmented Lagrangian constraint enforces the target sparsity at either layer or KV-head granularity.
We evaluate ConSA on two LLMs at the 0.6B and 1.7B scales.
Learned allocations consistently outperform rule-based baselines, with KV-head-wise allocation yielding clear gains over layer-wise allocation.
The learned patterns place SWA in the bottom layers and concentrate FA into contiguous middle-layer blocks, diverging from evenly interleaved patterns in rule-based methods.
This structure persists across model scales, sparsity levels, and allocation granularities, revealing a fine-grained spectrum of intrinsic attention behaviors that underlies the learned allocation.

\end{abstract}

\section{Introduction}\label{sec.intro}
Large language models have made attention cost a deployment bottleneck: full attention (FA) scales quadratically with sequence length in compute and linearly in KV cache~\cite{KwonLZ0ZY0ZS23}. 
Among efficient alternatives, sliding-window attention (SWA)~\cite{abs-2004-05150} restricts each token to a fixed local window, reducing both attention compute and per-head KV cache during inference.
However, the fixed window discards long-range dependencies, which can hurt tasks that require global context~\cite{XiaoZ0XLZ0024}.
A natural strategy to balance cost and capability is to combine FA and SWA within a single architecture.

Production models such as Mistral~\cite{abs-2310-06825}, Gemma~2~\cite{abs-2408-00118}, and MiMo-V2-Flash~\cite{MiMo-V2} have adopted such hybrid designs through hand-crafted interleaved patterns.
However, these manually specified allocations do not account for the heterogeneous attention behaviors across layers and heads in the original model~\cite{XiaoTZGYTF025}.
LoZA~\cite{abs-2512-23966} replaces manual design with a lightweight calibration stage that scores layers via learnable scalar weights and converts low-scoring layers to local attention.
Yet when calibrated on a small amount of pre-training data, such scalar scores may have limited discriminative power across layers, making layer selection less reliable under different target sparsity levels.
Moreover, prior work~\cite{XiaoTZGYTF025,abs-2512-23966,abs-2603-26380} on hybrid attention offers little analysis of what FA/SWA patterns emerge across layers and heads, and how these patterns relate to the intrinsic attention behaviors of the original model.

These limitations motivate a learnable allocation method that optimizes FA/SWA assignments under an explicit sparsity objective, while also calling for a finer-grained analysis of the intrinsic attention behaviors underlying the learned allocation.
We propose \textbf{\aname} (\textbf{Con}trollable \textbf{S}parsity in Hybrid \textbf{A}ttention), a framework for learning hybrid FA/SWA allocation under controllable sparsity.
Given a pre-trained Transformer and a user-specified target sparsity $\rho$, \aname formulates hybrid attention as a sparsity-constrained optimization problem: each attention unit receives a binary mask, parameterized by the hard concrete distribution under L0 regularization~\cite{louizos2018learning,XiaGZ024}, that selects between FA and SWA.
An augmented Lagrangian constraint is designed to enforce the target $\rho$, enabling the model to discover optimal allocations at either layer or KV-head granularity.
The mask parameters and model weights are first jointly optimized during a mask-learning stage, after which the learned masks are binarized and fixed for continued pre-training.

We further analyze the learned allocation patterns and the intrinsic attention behaviors of the models across model scales and sparsity levels.
The learned masks consistently place SWA in the bottom layers and concentrate FA into contiguous middle-layer blocks, diverging from the evenly interleaved patterns used in rule-based methods.
Examination of the attention behavior of representative layers and heads under the learned allocation reveals diverse attention spike ranges that extend beyond the retrieval-versus-streaming dichotomy described in prior work~\cite{XiaoTZGYTF025}, and align well with \aname's learned allocation.

Our contributions are threefold: (1)~We propose \aname, a framework that learns hybrid FA/SWA allocation via L0 regularization and augmented Lagrangian optimization at both layer-wise and KV-head-wise granularity, enabling users to specify an arbitrary target $\rho$ that is reliably satisfied during optimization. (2)~Experiments across two model scales (0.6B and 1.7B) and multiple sparsity levels show that learned allocations consistently outperform rule-based baselines, with KV-head-wise allocation yielding clear gains over layer-wise allocation. Ablation studies further confirm that the L0-Lagrangian formulation outperforms calibration-based approaches relying on unconstrained scalar gates with post-hoc ranking. (3)~Analysis of the learned patterns reveals a consistent SWA-bottom / FA-middle structure across model scales, sparsity levels, and allocation granularities. Examination of intrinsic attention behavior shows that this structure aligns with diverse attention spike ranges extending beyond the retrieval-versus-streaming dichotomy in prior work.

\section{Related Work}\label{sec.rw}
\paragraph{Efficient Attention Mechanisms.}
The quadratic scaling of full attention has led to a variety of efficient alternatives. 
Sliding-window attention (SWA)~\cite{abs-2004-05150,ZaheerGDAAOPRWY20,abs-1904-10509} is a common choice because it limits computational overhead and the KV cache footprint during inference. Other approaches include linear attention~\cite{KatharopoulosV020}, sparse attention with learned patterns~\cite{KitaevKL20,RoySVG21}, and state-space models~\cite{abs-2312-00752}. Our work does not introduce new attention mechanisms but focuses on how to distribute FA and SWA within a model. 

\paragraph{Hybrid Attention Architectures.}
Recent LLMs often combine FA and SWA via hand-crafted patterns: Mistral~\cite{abs-2310-06825} alternates SWA and FA layers, Gemma~2~\cite{abs-2408-00118} uses interleaving tied to scale, and Command-R and Jamba~\cite{LenzLABMPAAFPGJ25} adopt mixed types.
Two recent works move toward learned allocation: SwiAttn~\cite{Switch_Attention} routes tokens to FA or SWA via per-layer routers but must retain a unified KV cache; LoZA~\cite{abs-2512-23966} calibrates a per-layer scalar weight and converts the bottom-ranked layers to streaming sparse attention at a fixed $50\%$ ratio.
\aname differs by formulating FA/SWA allocation as a sparsity-constrained optimization problem, where an augmented Lagrangian constraint enforces a user-specified target; a detailed comparison is provided in Appendix~\ref{app.comparison}.

\paragraph{Attention Head Analysis.}
Attention heads are known to perform distinct roles, such as tracking position, syntax, or rare tokens~\cite{VoitaTMST19,ClarkKLM19}. 
More recent work identifies retrieval heads, which assign close attention mass to a few critical tokens across the full context, and streaming heads, which attend primarily to recent tokens and attention sinks; this classification is derived from output deviation on synthetic long-range retrieval tasks and has guided KV cache compression~\cite{XiaoTZGYTF025}.
\aname's learned allocation reveals a consistent SWA-bottom / FA-middle structure across model scales, sparsity levels, and allocation granularities. Analysis of representative layers and heads shows that their intrinsic attention spike ranges form a finer-grained spectrum beyond this binary classification, aligning well with the learned FA/SWA assignment.
\begin{figure*}[t]
    \centering
    \includegraphics[width=\textwidth]{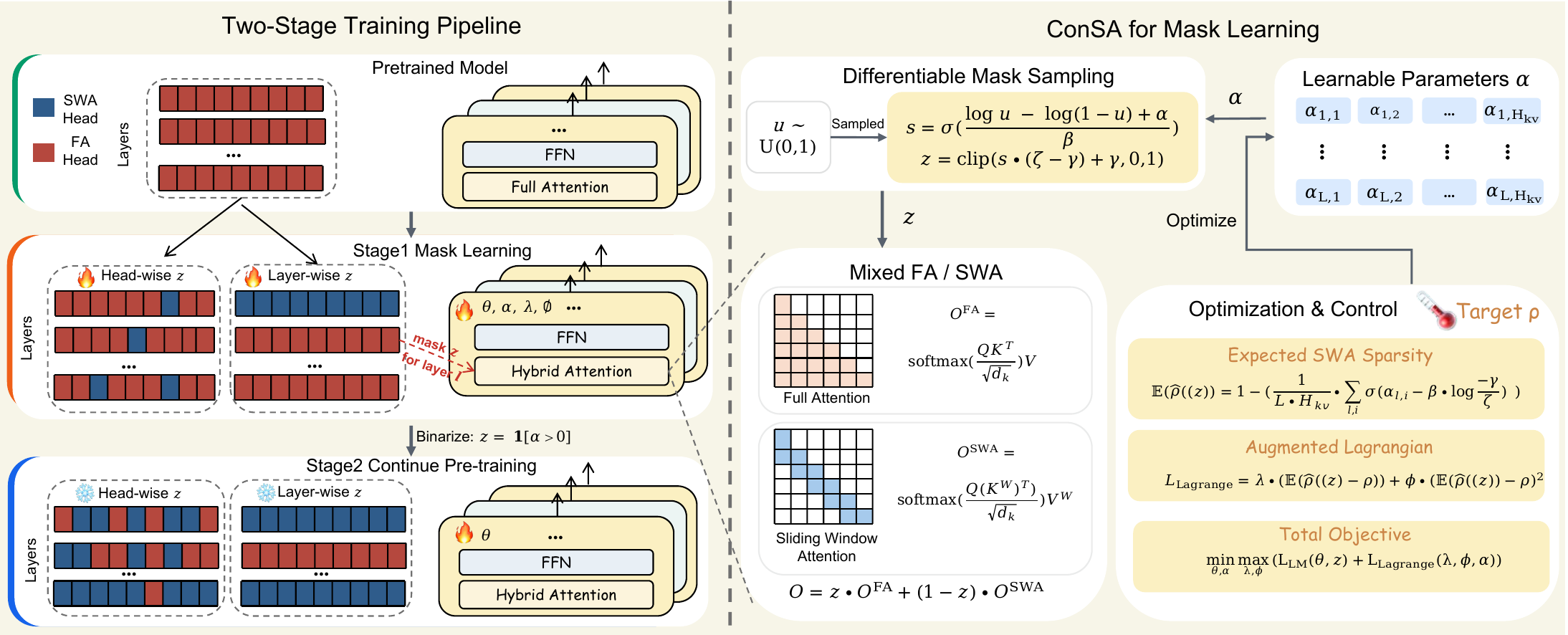}
    \caption{
    Overview of \aname.
    \textbf{Left:} the two-stage training pipeline. Stage~1 jointly optimizes the model parameters $\theta$, mask parameters $\alpha$, and Lagrange multipliers $\{\lambda, \phi\}$ on $1$B tokens, with the constraint $\hat{\rho}(z) = \rho$ enforcing the user-specified target sparsity. Stage~2 binarizes the masks and continues pre-training for $100$B tokens with a fixed FA/SWA assignment.
    \textbf{Right:} the per-head allocation mechanism. For each KV head $(l,i)$, a hard concrete mask $z_{l,i}$, parameterized by a learnable $\alpha_{l,i}$, selects between full attention (FA) and sliding-window attention (SWA).
    }
    \label{fig:method_overview}
\end{figure*}

\section{Preliminaries}\label{sec.bg}

We consider a Transformer with $L$ layers, each containing multiple key-value (KV) heads.
FA and SWA can be applied at different granularities; we formalize both at the KV-head level, which is the finest granularity considered in our method.

\paragraph{Full Attention (FA).}
The output of the $i$-th KV-head group in layer $l$ is:
\begin{equation}
\mathbf{O}_{l,i}^{\mathrm{FA}} = \mathrm{softmax}\!\left(\frac{\mathbf{Q}_{l,i} \mathbf{K}_{l,i}^\top}{\sqrt{d_k}}\right) \mathbf{V}_{l,i},
\label{eq:fa_output}
\end{equation}
where $\mathbf{Q}_{l,i}$ denotes the concatenated queries from all $g$ query heads in the group, and $\mathbf{K}_{l,i}, \mathbf{V}_{l,i} \in \mathbb{R}^{n \times d_k}$ are the shared key and value matrices for a sequence of length $n$ with head dimension $d_k$.
Under causal masking, FA allows each token to attend to all preceding tokens, incurring $O(n^2)$ compute and $O(n)$ KV cache per head group.

\paragraph{Sliding-Window Attention (SWA).}
SWA restricts each token to attend only to the $w$ most recent preceding tokens, where $w$ is a fixed window size:
\begin{equation}
\mathbf{O}_{l,i}^{\text{SWA}} = \mathrm{softmax}\!\left(\frac{\mathbf{Q}_{l,i} (\mathbf{K}_{l,i}^{w})^\top}{\sqrt{d_k}}\right) \mathbf{V}_{l,i}^{w},
\label{eq:swa_output}
\end{equation}
where $\mathbf{K}_{l,i}^{w}, \mathbf{V}_{l,i}^{w} \in \mathbb{R}^{w \times d_k}$ are the shared key and value matrices containing only the entries within the window.
With $w \ll n$, the KV cache is reduced from $O(n)$ to $O(w)$ and the compute cost from $O(n^2)$ to $O(nw)$ per head group, yielding substantial efficiency gains over FA.

\section{Method}\label{sec.method}

\subsection{Problem Formulation}
\label{sec:PF}

\aname formulates the design of hybrid attention as a sparsity-constrained allocation problem: given a pre-trained Transformer, the objective is to determine, for each KV head, whether it should perform full attention (FA) or sliding-window attention (SWA), such that the resulting hybrid model satisfies a user-specified target sparsity while preserving language modeling performance.

Let $\rho \in [0,1]$ denote the target sparsity ratio, defined as the fraction of KV heads assigned to SWA.
For the $i$-th KV head in layer $l$, we introduce a binary allocation variable $z_{l,i} \in \{0,1\}$ that selects between the two attention types.
The output of each KV-head group is then a hard selection between the two:
\begin{equation}
\hat{\mathbf{O}}_{l,i}
= z_{l,i} \cdot \mathbf{O}_{l,i}^{\mathrm{FA}}
+ (1-z_{l,i}) \cdot \mathbf{O}_{l,i}^{\mathrm{SWA}}.
\end{equation}

\aname applies this formulation at two levels of granularity. The \emph{head-wise} variant treats each $z_{l,i}$ as an independent variable, allowing different KV heads within the same layer to adopt different attention types. The \emph{layer-wise} variant constrains all KV heads in a layer to share a single allocation variable, $z_{l,i} = z_l$ for all $i$, which reduces the size of the search space.
The induced sparsity ratio $\hat{\rho}(z)$ under head-wise allocation is
\begin{equation}
\hat{\rho}(z) =
\hat{\rho}_{\mathrm{head}}(z) = 1 - \frac{1}{L \cdot H_{\mathrm{KV}}} \sum_{l=1}^{L} \sum_{i=1}^{H_{\mathrm{KV}}} z_{l,i},
\label{eq:rho-head}
\end{equation}
and under layer-wise allocation, it simplifies to
\begin{equation}
\hat{\rho}(z) =
\hat{\rho}_{\mathrm{layer}}(z) = 1 - \frac{1}{L} \sum_{l=1}^{L} z_l,
\label{eq:rho-layer}
\end{equation}
where $L$ is the number of layers and $H_{\mathrm{KV}}$ is the number of KV heads per layer.
The overall optimization problem is
\begin{equation}
\min_{\theta, z} \; \mathcal{L}_{\mathrm{LM}}(\theta, z)
\quad \text{s.t.} \quad \hat{\rho}(z) = \rho,
\label{eq:problem}
\end{equation}
where $\theta$ denotes the model parameters and $\mathcal{L}_{\mathrm{LM}}$ the autoregressive language modeling loss.

% 第二部分
\subsection{Differentiable Mask Learning with Hard Concrete}
\label{sec:LMLR}

Since $z_{l,i} \in \{0,1\}$ is binary, Eq.~\ref{eq:problem} is non-differentiable and cannot be optimized directly with gradients.
To jointly train $\theta$ and $z$, we parameterize each $z_{l,i}$ with the hard concrete distribution~\citep{louizos2018learning}, which assigns non-zero probability mass to $0$ and $1$ while remaining continuous and differentiable in between.
We refer to the resulting $z_{l,i}$ as a learnable binary mask.
Each mask is controlled by a learnable parameter $\alpha_{l,i} \in \mathbb{R}$ and is sampled as
\begin{equation}
\begin{aligned}
u &\sim \mathcal{U}(0,1), \\
s &= \sigma\!\left(
\tfrac{1}{\beta}\bigl(\log u - \log(1-u) + \alpha_{l,i}\bigr)
\right), \\
\bar{s} &= s \cdot (\zeta-\gamma) + \gamma, \\
z_{l,i} &= \min\bigl(1,\, \max(0,\, \bar{s})\bigr),
\end{aligned}
\end{equation}
where $\sigma$ is the sigmoid function, $\beta$ the temperature, and $\zeta > 1$ and $\gamma < 0$ the stretch parameters.

During training, $z_{l,i}$ is sampled stochastically and gradients with respect to $\alpha_{l,i}$ are obtained by reparameterizing the noise variable $u$.
We initialize all $\alpha_{l,i}$ to $5.0$, which makes the expected mask value $\bar{z}_{l,i}$ close to $1$, so every head starts as FA at the beginning of training.
This warm-start anchors mask learning at the original full-attention configuration of the pre-trained model and lets the optimizer turn a head into SWA only when doing so does not hurt the loss.

% 第三部分
\subsection{Lagrangian Optimization under Sparsity Constraints}
\label{sec:LRSC}

The problem in Eq.~\ref{eq:problem} is a constrained optimization that requires the realized sparsity to match the target $\rho$.
We address it through augmented Lagrangian relaxation, which converts the equality constraint into an additive penalty and yields the unified training objective
\begin{equation}
\resizebox{0.48\textwidth}{!}{
$
\displaystyle
\min_{\theta, \alpha} \; \max_{\lambda, \phi} \;
\mathcal{L}_{\mathrm{LM}}(\theta, z)
+ \lambda \cdot \bigl(\hat{\rho}(z)-\rho\bigr)
+ \phi \cdot \bigl(\hat{\rho}(z)-\rho\bigr)^2
$
}
\label{eq:total-loss}
\end{equation}
where $\lambda$ is a Lagrange multiplier and $\phi$ is an adaptive quadratic coefficient. The linear term pushes the realized sparsity toward $\rho$, while the quadratic term penalizes larger violations and stabilizes the optimization dynamics around the constraint.
Under this min-max formulation, $\theta$ and $\alpha$ are updated by gradient descent, while $\lambda$ and $\phi$ are updated by gradient ascent.

As a stochastic sample from the hard concrete distribution, $z$ fluctuates across forward passes, which produces high-variance gradients with respect to $\alpha$.
We therefore constrain its expectation $\mathbb{E}[\hat{\rho}(z)]$ instead, which is a smooth, deterministic function of $\alpha$.
By the linearity of expectation,
\begin{equation}
\begin{aligned}
\mathbb{E}[\hat{\rho}(z)] &= 1 - \frac{1}{L \cdot H_{\mathrm{KV}}} \sum_{l=1}^{L} \sum_{i=1}^{H_{\mathrm{KV}}} \mathbb{E}[z_{l,i}], \\
\mathbb{E}[z_{l,i}] &= 1 - F_{\bar{s}_{l,i}}(0 \mid \alpha_{l,i}) \\
&= \sigma\!\left(\alpha_{l,i} - \beta \log \tfrac{-\gamma}{\zeta}\right),
\end{aligned}
\label{eq:expected_sparsity}
\end{equation}
where $F_{\bar{s}_{l,i}}(\cdot \mid \alpha_{l,i})$ is the cumulative distribution function of the stretched concrete variable $\bar{s}_{l,i}$, and $\sigma$ is the sigmoid function.
Replacing $\hat{\rho}(z)$ in Eq.~\ref{eq:total-loss} with $\mathbb{E}[\hat{\rho}(z)]$ gives the objective actually optimized during training:
\begin{equation}
\resizebox{0.48\textwidth}{!}{
$
\min_{\theta, \alpha} \; \max_{\lambda, \phi} \;
\mathcal{L}_{\mathrm{LM}}(\theta, z)
+ \lambda \cdot \bigl(\mathbb{E}[\hat{\rho}(z)]-\rho\bigr)
+ \phi \cdot \bigl(\mathbb{E}[\hat{\rho}(z)]-\rho\bigr)^2.
$
}
\label{eq:total-loss-final}
\end{equation}
This substitution is exact in the limit: by a property of the hard concrete distribution, each $\mathbb{E}[z_{l,i}]$ concentrates on $\{0,1\}$ as training proceeds, so the expected sparsity converges to the realized sparsity $\hat{\rho}(z)$ at convergence.

After the augmented Lagrangian penalty converges to zero, we drop the stochastic sampling and binarize each learned mask as
\begin{equation}
z_{l,i} = \mathbbm{1}[\alpha_{l,i} > 0].
\end{equation}
The binarized masks give a fixed FA/SWA assignment that is used in all subsequent forward passes.
The full training pipeline is described in Section~\ref{sec:exp.setup}.
% The full training pipeline, including how mask learning is followed by continued pre-training under the binarized configuration, is described in Section~\ref{sec:exp.setup}.

\section{Experiments}\label{sec.exp}

\subsection{Experimental Setup} 
\label{sec:exp.setup}

\begin{table*}[t]
\centering
\renewcommand{\arraystretch}{1.05}
\resizebox{\textwidth}{!}{%
\begin{tabular}{lcccccc}
\toprule
\textbf{Method} 
& \textbf{MMLU} 
& \textbf{LogiQA-EN} 
& \textbf{LogiQA-CN} 
& \textbf{CSQA} 
& \textbf{PIQA} 
& \textbf{SIQA} \\
\midrule
Dense FA
& 45.51 & 34.31 & \underline{33.23} & 50.04 & 56.91 & \textbf{54.86} \\
\midrule
\rowcolor{customgray!10}
\aname (\emph{head-wise, single-layer})
& \textbf{45.76} & \textbf{36.92} & \textbf{34.92} & \underline{52.09} & \textbf{61.32} & 53.94 \\
\rowcolor{customgray!10}
\aname (\emph{head-wise, all-layers})
& \underline{45.55} & \underline{35.08} & 33.08 & 51.27 & 57.78 & \underline{54.40} \\
Rule (\emph{head-wise})
& 45.15 & 34.62 & 32.46 & 49.22 & \underline{58.76} & 53.28 \\
\midrule
\rowcolor{customgray!10}
\aname (\emph{layer-wise})
& 45.45 & 32.92 & 31.54 & \textbf{52.99} & 56.47 & 53.89 \\
Rule (\emph{layer-wise}) 
& 44.03 & 31.71 & 31.02 & 51.43 & 56.12 & 52.30 \\
\midrule
\midrule
\textbf{Method} 
& \textbf{ARC-C} 
& \textbf{Hella} 
& \textbf{ARC-E} 
& \textbf{WebQA-CN} 
& \textbf{CN-GEN}
& \textbf{Average} \\
\midrule
Dense FA
& 51.02 & 36.35 & 69.91 & 54.58 & \textbf{39.38} & 47.83 \\
\midrule
\rowcolor{customgray!10}
\aname (\emph{head-wise, single-layer})
& 51.71 & \textbf{37.93} & 71.00 & \textbf{57.15} & 36.89 & \textbf{49.06} \\
\rowcolor{customgray!10}
\aname (\emph{head-wise, all-layers})
& \textbf{52.05} & 34.27 & \underline{71.21} & \underline{56.86} & \underline{37.88} & \underline{48.13} \\
Rule (\emph{head-wise})
& 51.19 & 34.61 & 69.11 & 56.16 & 35.87 & 47.31 \\
\midrule
\rowcolor{customgray!10}
\aname (\emph{layer-wise})
& \underline{51.79} & \underline{36.98} & \textbf{71.30} & 55.29 & 36.51 & 47.74 \\
Rule (\emph{layer-wise}) 
& 50.43 & 34.31 & 67.80 & 55.41 & 36.39 & 46.45 \\
\bottomrule
\end{tabular}%
}
\caption{
Comparison of head-wise and layer-wise FA/SWA allocation on 1.7B at target sparsity $\rho = 0.50$.
\aname and the rule-based baselines are matched at the same $\rho$, while FA ($\rho = 0$) serves as the dense reference.
The best results are in \textbf{bold}, and the second-best results are \underline{underlined}.
}
\label{tab:main_result}
\vspace{-2mm}
\end{table*}

\paragraph{Training Pipeline.}
We train the model in two stages. \emph{Stage~1 (Mask Learning, 1B tokens)} jointly optimizes the model parameters $\theta$, mask parameters $\alpha$, and Lagrange multipliers $\lambda$ and $\phi$, starting from a pre-trained checkpoint. 
In this stage, masks are sampled from the hard concrete distribution with a constraint that drives the expected sparsity $\mathbb{E}[\hat{\rho}(z)]$ toward the target $\rho$. 
In \emph{Stage~2 (Continued Pre-training, 100B tokens)}, we binarize the masks via $z_{l,i} = \mathbbm{1}[\alpha_{l,i} > 0]$ and continue training on the resulting fixed FA/SWA assignments to let the weights adapt to the new configuration.
(see Appendix~\ref{app.Training Details} for further details).

\paragraph{Models.}
We pre-train two dense Transformer LLMs from scratch to evaluate \aname: a 0.6B-parameter model and a 1.7B-parameter model. 
Both adopt a standard GQA architecture with 28 layers, 16 query heads, and 8 KV heads per layer, differing only in the hidden dimension (Table~\ref{tab:model_para}).
We run the main downstream evaluation on 1.7B (Table~\ref{tab:main_result}) and use 0.6B for ablation studies and pattern visualization, since its smaller size lets us sweep over more sparsity levels and granularities at a lower computational cost.

\paragraph{Baselines.}
We compare the six configurations listed in Table~\ref{tab:main_result}, all evaluated under matched continued pre-training:
1)~Dense FA, the full-attention reference with $\rho=0$ in which every KV head performs full attention;
2)~\aname (\emph{head-wise, single- layer}), the head-wise variant of \aname trained under a per-layer sparsity constraint that requires each layer to independently satisfy $1 - \frac{1}{H_{\mathrm{KV}}} \sum_{i} z_{l,i} = \rho$;
3)~\aname (\emph{head-wise, all-layers}), the head-wise variant trained under the global constraint in Eq.~\ref{eq:rho-head}, where $\rho$ is imposed only on the full pool of $L \cdot H_{\mathrm{KV}}$ KV heads, so that the optimizer can distribute the SWA budget unevenly across layers;
4)~Rule (\emph{head-wise}), a static head-wise pattern with a hand-crafted SWA/FA assignment within each layer at the target sparsity $\rho$;
5)~\aname (\emph{layer-wise}), the layer-wise variant of \aname trained under the constraint in Eq.~\ref{eq:rho-layer}, in which all KV heads within a layer share a single allocation $z_l$;
6)~Rule (\emph{layer-wise}), a static layer-wise interleaving in the style of Mistral~\cite{abs-2310-06825} and Gemma~2~\cite{abs-2408-00118}, where SWA and FA layers alternate at the same $\rho$.

\paragraph{Evaluation.}
We evaluate our models on a range of English and Chinese benchmarks covering knowledge and reasoning. General knowledge is measured via MMLU~\cite{MMLU}, while logical reasoning is assessed using LogiQA-EN and LogiQA-CN~\cite{logiqa}. For commonsense reasoning, we include CommonsenseQA (CSQA)~\cite{CSQA}, PIQA~\cite{piqa}, and SocialIQA (SIQA)~\cite{siqa}. Scientific and contextual reasoning are tested using ARC-Challenge (ARC-C), ARC-Easy (ARC-E)~\cite{arc}, and HellaSwag (Hella)~\cite{hella}, alongside open-domain question answering with WebQA-CN~\cite{webqa-cn}. The evaluation also covers two Chinese generation tasks (CN-GEN): scientific summarization from CSL~\cite{CSL} and story generation from LOT~\cite{LOT}.

% 实验结果部分
\subsection{Main Results}

\paragraph{Learned Allocation Outperforms Rule Allocation.}
Tab.~\ref{tab:main_result} compares \aname against rule-based baselines on 1.7B at $\rho = 0.50$.
Under head-wise allocation, the single-layer variant of \aname yields a 3.7\% relative improvement in average accuracy over Rule (\emph{head-wise}), with consistent gains across all eleven benchmarks.
Under layer-wise allocation, \aname likewise outperforms Rule (\emph{layer-wise}) by a 2.8\% relative margin on average.
The consistent advantage observed under both granularities indicates that the allocation learned by \aname captures FA/SWA configurations that are unreachable through hand-crafted interleaving.

\paragraph{Head-wise \aname Matches or Exceeds Dense FA.}
Although \aname operates at $\rho = 0.50$ while Dense FA uses no sparsity ($\rho = 0$), the head-wise (single-layer) variant surpasses Dense FA by a 2.6\% relative margin on average.
The main exception is CN-GEN, where performance drops due to long-range dependencies that fall outside the SWA window.
On the remaining benchmarks, sequence lengths generally fall within the SWA window, so the selected heads attend to the same context as FA heads at test time.
This indicates that the performance gap originates from training, where local attention on these heads may act as an implicit regularizer that encourages more focused attention patterns in the learned weights.

\begin{figure}[t]
    \centering
    \includegraphics[width=\columnwidth]{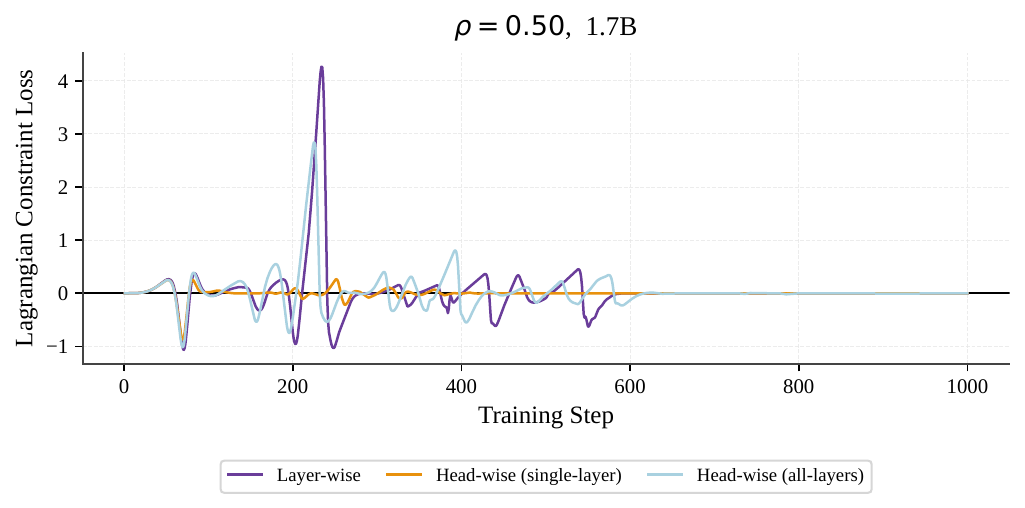}
    \caption{
    Convergence of the Lagrangian constraint loss during Stage~1 mask learning on 1.7B at $\rho = 0.50$ under three allocation granularities.
    }
    \label{fig:lagrangian_convergence}
\end{figure}

\paragraph{Head-wise Granularity Drives Most of the Improvement.}
Both head-wise variants of \aname outperform the layer-wise variant despite sharing the same training framework and target $\rho$, indicating that the granularity of allocation is a key factor.
Among the two head-wise variants, the single-layer variant outperforms the all-layers variant.
We attribute this to the difference in constraint structure: the all-layers variant imposes $\rho$ as a single global constraint over all $L \times H_{\mathrm{KV}}$ heads, resulting in a substantially larger search space that makes the Lagrangian constraint harder to satisfy during mask learning.
As shown in Figures~\ref{fig:lagrangian_convergence} and~\ref{fig:lagrangian_06b_050}, the sparsity loss of the all-layers variant converges more slowly than that of the single-layer variant, which enforces $\rho$ independently at each layer.
The per-layer constraint, therefore, acts as a structural prior that regularizes the optimization and yields a more effective allocation.

% 拉格朗日loss图

% 消融实验图
\begin{figure}[t]
    \centering
    \begin{subfigure}[b]{\columnwidth}
        \centering
        \includegraphics[width=\textwidth]{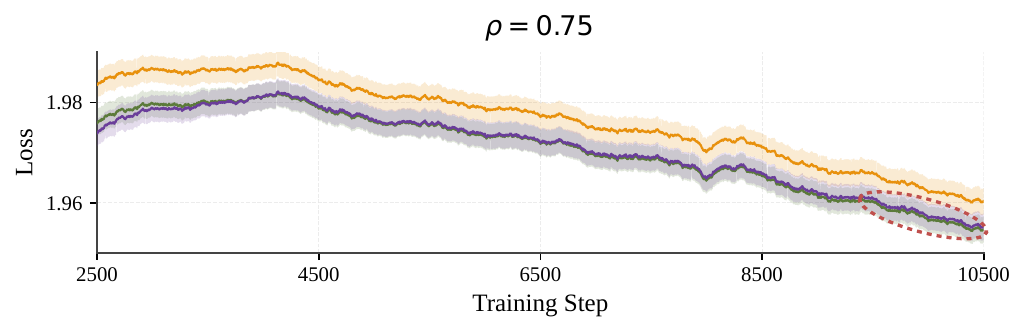}
    \end{subfigure}
    \vspace{0.5em}
    \begin{subfigure}[b]{\columnwidth}
        \centering
        \includegraphics[width=\textwidth]{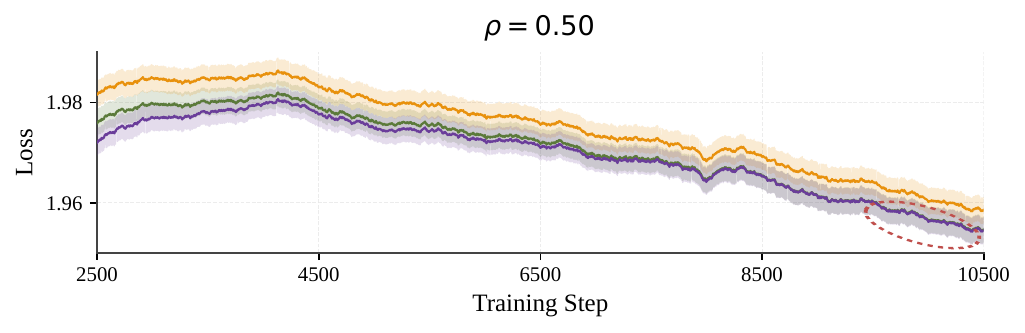}
    \end{subfigure}
    \vspace{0.5em}
    \begin{subfigure}[b]{\columnwidth}
        \centering
        \includegraphics[width=\textwidth]{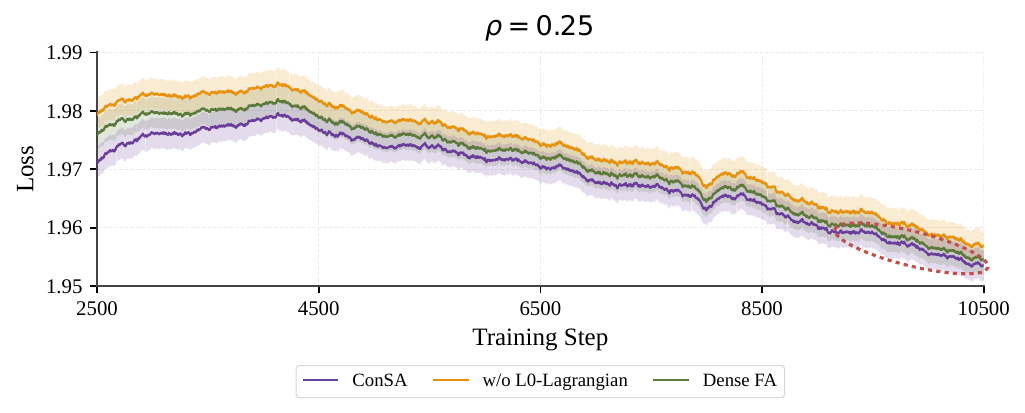}
    \end{subfigure}
    \caption{
    Training loss trajectories on 0.6B under layer-wise allocation at $\rho \in \{0.25, 0.50, 0.75\}$. 
    Each panel compares \aname, the ablation variant (w/o L0-Lagrangian), and Dense FA over 20B tokens of continued pre-training. 
    \textcolor{customred}{Dashed circles} highlight the final convergence region where the relative ordering of \aname and Dense FA shifts across sparsity levels.
    }
    \label{fig:ablation_loss}
\end{figure}

\subsection{Convergence of the Lagrangian Constraint}
\label{sec:convergence}
To verify that the learned masks meet the target sparsity, we monitor the Lagrangian constraint loss, ${\mathcal{L}}_{\text{Lagrange}}=\lambda \cdot (\mathbb{E}[\hat{\rho}(z)]-\rho) + \phi \cdot (\mathbb{E}[\hat{\rho}(z)]-\rho)^2$, during the 1B-token Stage~1 mask-learning phase.
Figure~\ref{fig:lagrangian_convergence} reports the trajectory of ${\mathcal{L}}_{\text{Lagrange}}$ for the 1.7B model at $\rho = 0.50$ across the three allocation granularities.

\paragraph{All Granularities Converge within the 1B-Token Budget.}
Under the min-max formulation of the augmented Lagrangian, the Lagrange multipliers and mask parameters compete before reaching equilibrium.
As a result, the constraint loss does not decrease monotonically but instead oscillates, converging to zero only when the constraint $\hat{\rho}(z) = \rho$ is satisfied.
Despite these transient oscillations, all three configurations drive the constraint loss to near zero within 1,000 training steps, confirming that the target sparsity can be reliably achieved within the 1B-token mask-learning budget.
The effect of granularity on convergence speed and its connection to downstream performance are analyzed in Appendix~\ref{appdx:lagrangian-convergence}.

\subsection{Ablation Study}
\label{sec:ablation}
\paragraph{Setup.}
To isolate the contribution of the L0-Lagrangian formulation in \aname, we construct an ablation variant that removes it and instead adopts a calibration-based strategy analogous to those used in LoZA~\cite{abs-2512-23966} and DuoAttention~\cite{XiaoTZGYTF025}.
Each attention unit is equipped with an unconstrained scalar gate $\alpha_i$, initialized to 1.0 and clamped to $[0, 1]$ during training. The attention output at layer $l$, head $i$ is computed as:
\begin{equation}
\hat{\mathbf{O}}_{l,i}
= \alpha_{l,i} \cdot \mathbf{O}_{l,i}^{\mathrm{FA}}
+ (1-\alpha_{l,i}) \cdot \mathbf{O}_{l,i}^{\mathrm{SWA}}.
\end{equation}
The training objective is:
\begin{equation}
\begin{aligned}
\mathcal{L} = \mathcal{L}_{\mathrm{LM}} + \lambda \cdot \mathcal{L}_{\mathrm{L1}}, \\
\mathcal{L}_{\mathrm{L1}} = \frac{1}{N}\sum_i |\alpha_i|,
\end{aligned}
\end{equation}
where $\lambda = 0.05$ following DuoAttention.
Since a higher $\alpha_{l,i}$ indicates a stronger preference for full attention, the final FA/SWA assignment is obtained by sorting all gates in ascending order and assigning the bottom-$\rho$ fraction to SWA, following the ranking-based selection of LoZA.
We compare \aname with this ablation and the Dense FA baseline on 0.6B under layer-wise allocation at $\rho \in \{0.25, 0.50, 0.75\}$.
Both \aname and the ablation train masks on 1B tokens, followed by 20B tokens of continued pre-training under the resulting fixed configuration.
We report the loss trajectory over the 20B-token stage in Figure~\ref{fig:ablation_loss}.
The detailed training setup is provided in Appendix~\ref{appdx:ablation-detail}.

\paragraph{ConSA vs.\ Ablation Variant.}
As shown in Figure~\ref{fig:ablation_loss}, \aname achieves a lower final loss than the ablation variant at every sparsity level, with the gap emerging early in training and persisting throughout.
By binding the mask distribution directly to the target $\rho$ during optimization, the L0-Lagrangian formulation steers the mask toward configurations that post-hoc selection based on unconstrained scalar weights cannot reach.
The gap between \aname and the ablation widens as $\rho$ increases from 0.25 to 0.75, suggesting that the Lagrangian constraint becomes increasingly important at higher sparsity levels where the optimization landscape grows more challenging.

% 模式分析部分
\begin{figure}[t]
    \centering
    \begin{subfigure}[b]{\columnwidth}
        \centering
        \includegraphics[width=\textwidth]{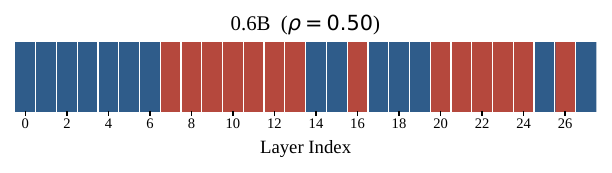}
    \end{subfigure}
    \vspace{0.5em}
    \begin{subfigure}[b]{\columnwidth}
        \centering
        \includegraphics[width=\textwidth]{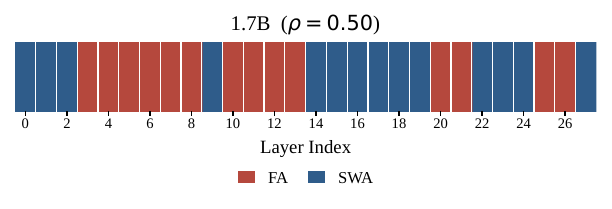}    \end{subfigure}
    \caption{Learned layer-wise FA/SWA allocation at $\rho = 0.50$. 
    Each cell indicates whether a layer uses \textcolor{customred}{FA (red)} or \textcolor{customblue}{SWA (blue)}.
    }
    \label{fig:layer_patterns}
\end{figure}

\begin{figure}[t]
    \centering
    \includegraphics[width=\columnwidth]{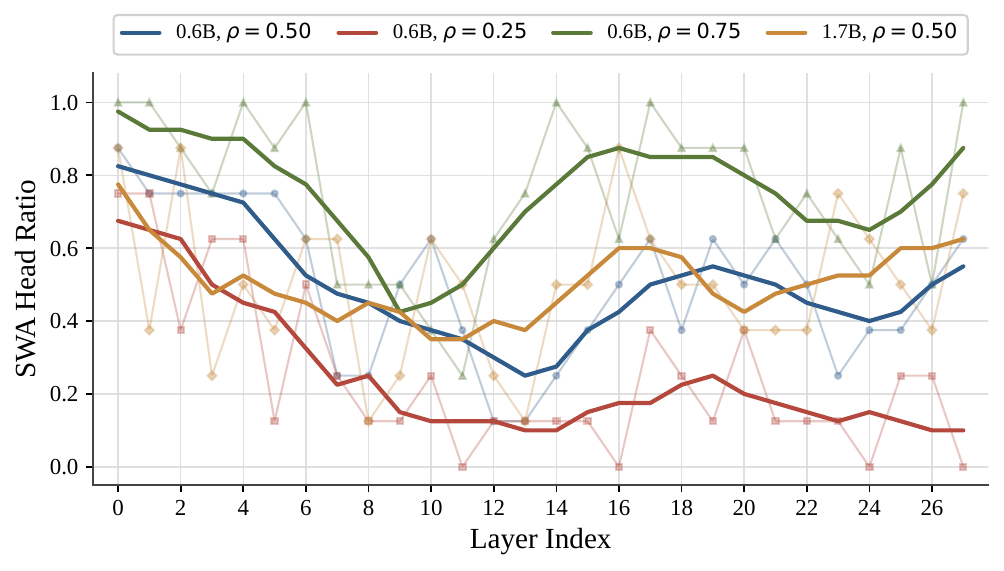}
    \caption{
    Per-layer SWA head ratios under head-wise allocation across four configurations. 
    The bold curves show moving averages, computed from the raw per-layer ratios shown by the faded curves.
    }
    \label{fig:head_ratio}
\end{figure}

% case study图
\begin{figure*}[t]
    \centering
   \begin{subfigure}[b]{0.32\textwidth}
    \centering
    \includegraphics[width=\textwidth]{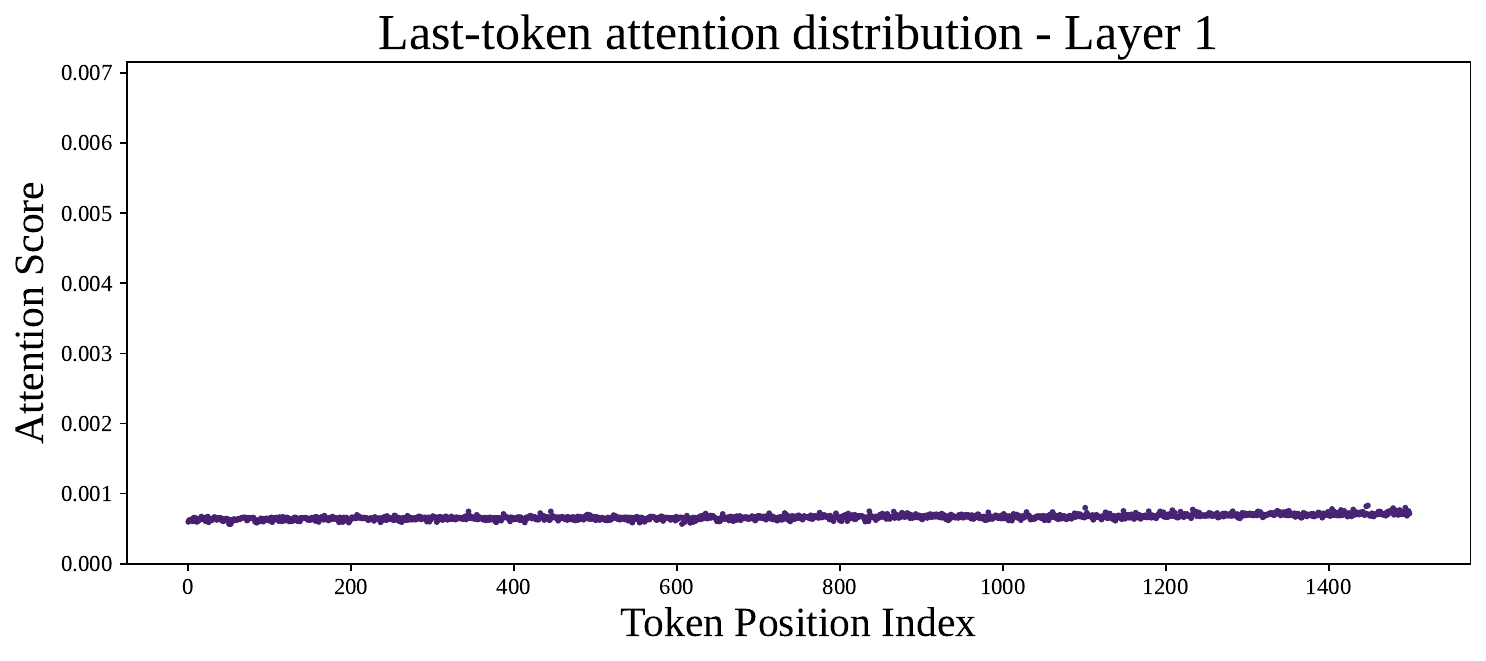}
    \caption{Uniform}
    \label{fig:case_L01}
\end{subfigure}
\hfill
\begin{subfigure}[b]{0.32\textwidth}
    \centering
    \includegraphics[width=\textwidth]{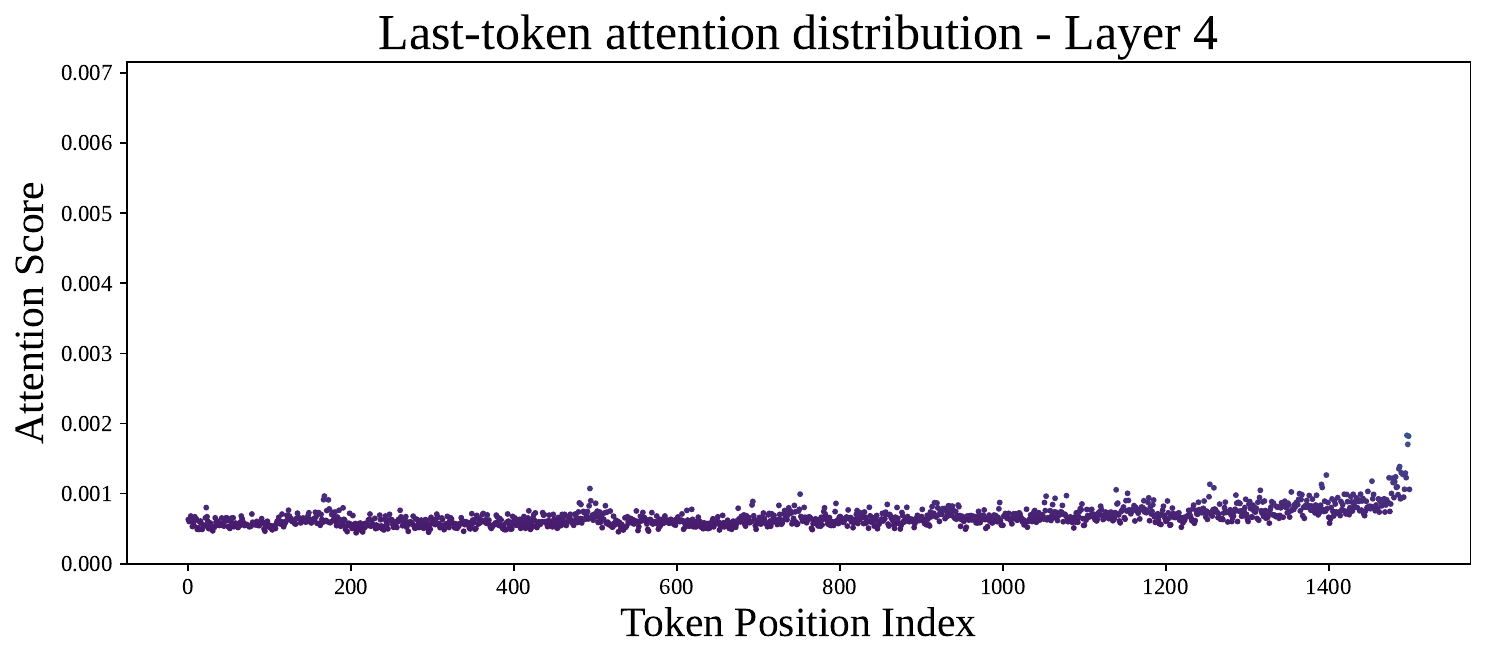}
    \caption{Weakly local}
    \label{fig:case_L04}
\end{subfigure}
\hfill
\begin{subfigure}[b]{0.32\textwidth}
    \centering
    \includegraphics[width=\textwidth]{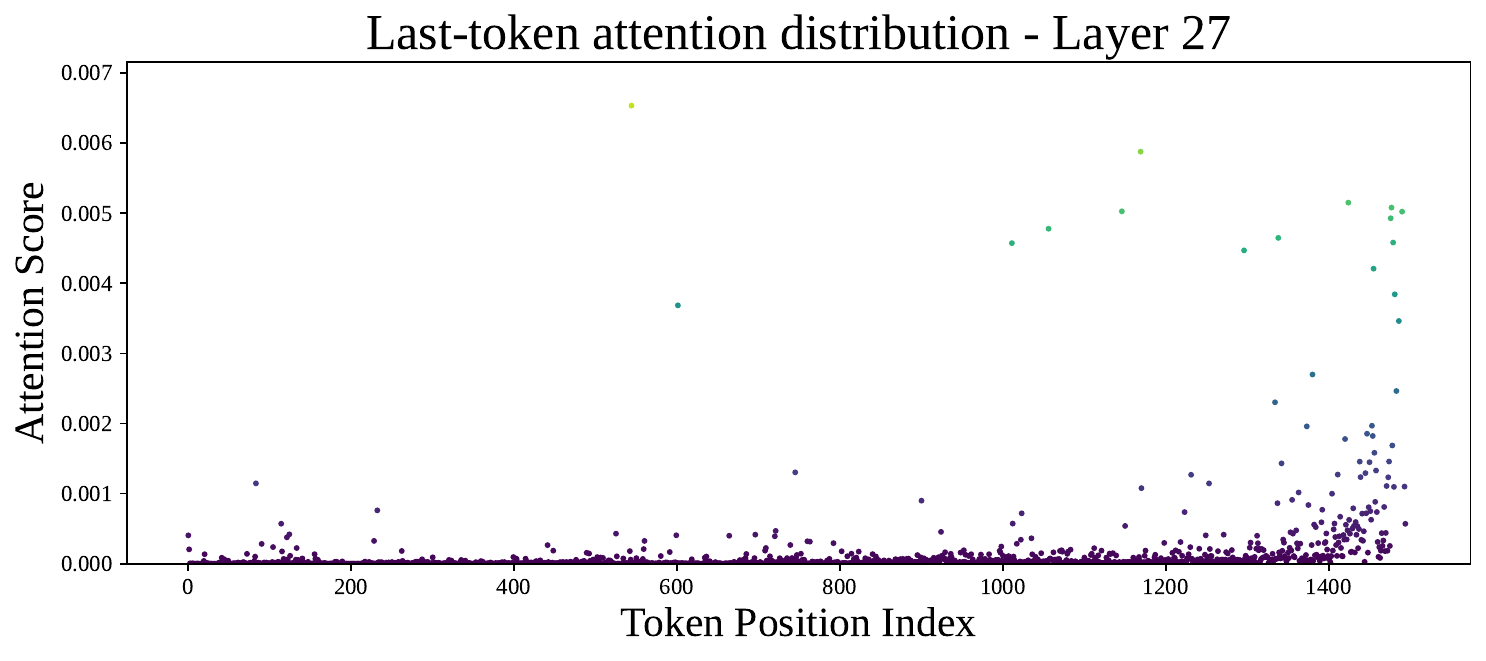}
    \caption{Strongly local}
    \label{fig:case_L27}
\end{subfigure}

% \vspace{0.4em}

\begin{subfigure}[b]{0.32\textwidth}
    \centering
    \includegraphics[width=\textwidth]{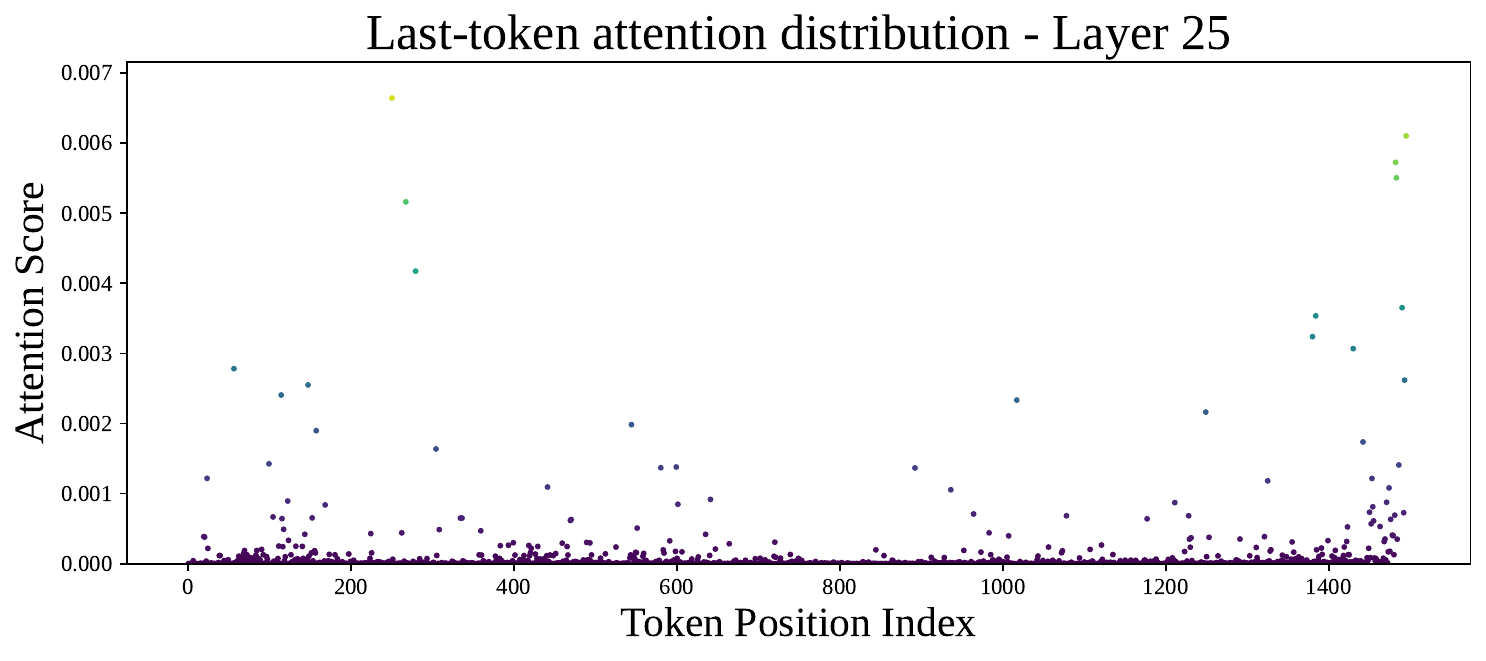}
    \caption{Sparse broad}
    \label{fig:case_L25}
\end{subfigure}
\hfill
\begin{subfigure}[b]{0.32\textwidth}
    \centering
    \includegraphics[width=\textwidth]{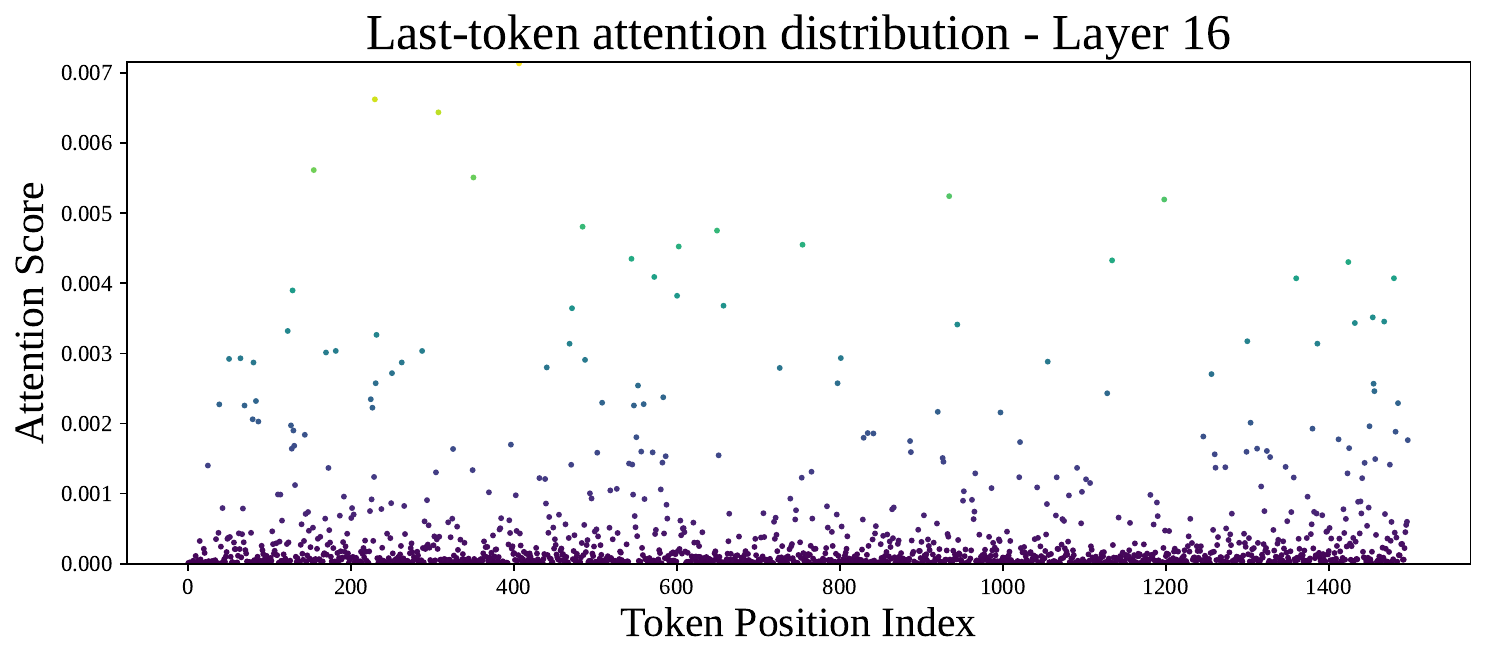}
    \caption{Dense broad}
    \label{fig:case_L16}
\end{subfigure}
\hfill
\begin{subfigure}[b]{0.32\textwidth}
    \centering
    \includegraphics[width=\textwidth]{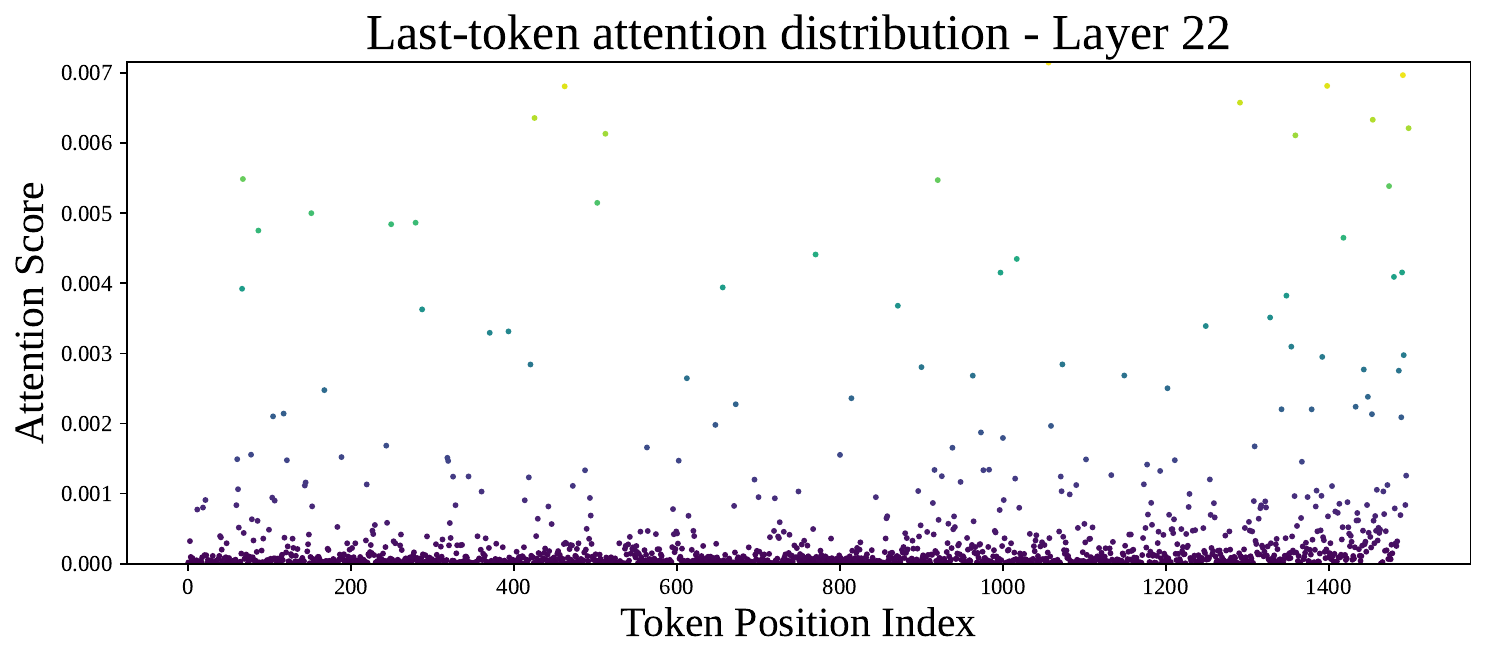}
    \caption{Dense broad}
    \label{fig:case_L22}
\end{subfigure}

  \caption{Last-token attention distribution across representative layers of the 0.6B model, spanning the spectrum from uniform to dense broad attention.}
    \label{fig:casestudy_main}
\end{figure*}

\subsection{Analysis of Learned Allocation Patterns}
\label{sec:pattern}

We analyze the learned FA/SWA allocation patterns across model scales, sparsity levels, and granularities to understand what configurations emerge from the L0-Lagrangian optimization.

\paragraph{Layer-wise Allocation Patterns.}
In contrast to rule-based approaches that interleave FA and SWA at a fixed ratio, the learned masks concentrate FA into contiguous middle-layer blocks, suggesting that adjacent FA layers working in concert are more beneficial than FA layers evenly spread across the network (Figure~\ref{fig:layer_patterns}).
Notably, the first layer is consistently assigned to SWA across all configurations, contradicting the design choice in several rule-based methods that designate the first layer as FA to capture global context early.
The concentration of FA in the middle layers is consistent with prior analyses showing that intermediate layers play a central role in semantic integration and reasoning~\cite{ClarkKLM19,VoitaTMST19,chen-etal-2025-improving-reasoning}, which may require a broader attention scope.
This structure is shared across both model scales (Figure~\ref{fig:layer_patterns}) and sparsity levels (Figure~\ref{fig:layerwise_025_075}), though the FA block shifts to slightly earlier layers in 1.7B.

\paragraph{Head-wise Allocation Patterns.}
Under head-wise allocation, the per-layer SWA head ratio follows an approximate W-shaped trend, with peaks at the bottom and top layers and dips in the middle (Figure~\ref{fig:head_ratio}).
This trend is shared across model scales and the higher sparsity levels ($\rho \in \{0.50, 0.75\}$), while at $\rho = 0.25$ the overall sparsity budget is too low for the top-layer recovery to manifest clearly.
The consistent dip in the middle layers aligns with the layer-wise observation that this region serves as the primary site for full attention.
Full per-head allocation heatmaps are provided in Appendix~\ref{appdendix:Patterns}.

\subsection{Analysis of Attention Behavior}
\label{sec:attention-analysis}

Across sparsity levels, \aname's learned allocation reveals that different layers exhibit distinct preferences for FA or SWA (Figure~\ref{fig:layerwise_025_075}).
To examine their intrinsic attention behaviors, we select six representative layers and visualize their last-token attention distribution, using the pre-trained checkpoint before Stage~1 mask learning.
For each layer, we average the attention scores of all heads into a layer-wise distribution.

\paragraph{Diverse Attention Spikes Ranges.}
Figure~\ref{fig:casestudy_main} reveals that these layers exhibit a fine-grained spectrum of attention patterns beyond the retrieval-versus-streaming dichotomy described in prior work~\cite{XiaoTZGYTF025}.
Among the layers assigned to SWA across all sparsity levels (L1, L4, L27), L1 displays a \emph{uniform} pattern (Figure~\ref{fig:case_L01}) with attention spread nearly evenly and at low magnitude across the full context;
L4 is \emph{weakly local} (Figure~\ref{fig:case_L04}), mostly uniform but with a mild rise in the last few hundred positions;
and L27 is \emph{strongly local}(Figure~\ref{fig:case_L27}), with attention sharply concentrated near the tail.
The layers assigned to FA across all sparsity levels, L16 and L22, show \emph{dense broad} attention pattern (Figures~\ref{fig:case_L16},~\ref{fig:case_L22}), with high-magnitude spikes spanning a wide range of the sequence.
The switching layer L25, which transitions from FA at $\rho = 0.25$ to SWA at $\rho \in \{0.50, 0.75\}$, shows \emph{sparse broad} attention pattern (Figure~\ref{fig:case_L25}): spikes appear at distant positions, but are confined to a small number of localized regions, placing it between the local and dense broad extremes.
This spectrum generally aligns with \aname's allocation: layers with narrower attention spike ranges tend to be assigned SWA first as the sparsity budget grows.
Further analysis is provided in Appendix~\ref{appendix:Attention Behavior}.

\section{Conclusion}\label{sec.con}
In this paper, we introduced \aname, a principled framework for learning hybrid FA/SWA attention configurations with controllable sparsity. 
By formulating FA/SWA allocation as a Lagrangian-constrained L0 optimization problem, \aname learns binary masks at both layer-wise and KV-head-wise granularities that meet user-specified sparsity targets. 
Analysis of the learned patterns reveals a counterintuitive yet architecture-consistent principle: bottom layers are predominantly assigned SWA, whereas FA concentrates in middle layers. 
This challenges prevailing intuitions about the necessity of global attention and offers practical utility and interpretive insights for the community.

\clearpage
\section*{Limitations}
First, the SWA window size is fixed throughout all experiments; jointly optimizing the window size and the FA/SWA allocation may yield further gains.
Second, although the cross-scale consistency of the identified patterns is promising, a more comprehensive evaluation across a broader spectrum of model scales and architectural designs is required to verify the generalizability of the proposed allocation principle.

\section*{Ethics Statement}
Our work examines architectural modifications to large language models, using publicly available pre-training corpora and evaluation benchmarks, all accessed under their respective open licenses for academic research use.
No personal or sensitive data is used during training or evaluation. 
The proposed method does not introduce new deployment risks beyond those commonly associated with language models.

% \section*{Acknowledgments}

\bibliography{reference}

@article{abs-2310-06825,
  author       = {Albert Q. Jiang and
                  Alexandre Sablayrolles and
                  Arthur Mensch and
                  Chris Bamford and
                  Devendra Singh Chaplot and
                  Diego de Las Casas and
                  Florian Bressand and
                  Gianna Lengyel and
                  Guillaume Lample and
                  Lucile Saulnier and
                  L{\'{e}}lio Renard Lavaud and
                  Marie{-}Anne Lachaux and
                  Pierre Stock and
                  Teven Le Scao and
                  Thibaut Lavril and
                  Thomas Wang and
                  Timoth{\'{e}}e Lacroix and
                  William El Sayed},
  title        = {Mistral 7B},
  journal      = {CoRR},
  volume       = {abs/2310.06825},
  year         = {2023},
  url          = {https://doi.org/10.48550/arXiv.2310.06825},
  doi          = {10.48550/ARXIV.2310.06825},
  eprinttype   = {arXiv},
  eprint       = {2310.06825},
  bibsource    = {dblp computer science bibliography, https://dblp.org}
}

@article{abs-2408-00118,
  author       = {Gemma Team},
  title        = {Gemma 2: Improving Open Language Models at a Practical Size},
  journal      = {CoRR},
  volume       = {abs/2408.00118},
  year         = {2024},
  url          = {https://doi.org/10.48550/arXiv.2408.00118},
  doi          = {10.48550/ARXIV.2408.00118},
  eprinttype   = {arXiv},
  eprint       = {2408.00118},
  bibsource    = {dblp computer science bibliography, https://dblp.org}
}

@article{abs-2512-23966,
  author       = {Chen Zhang and
                  Yang Bai and
                  Jiahuan Li and
                  Anchun Gui and
                  Keheng Wang and
                  Feifan Liu and
                  Guanyu Wu and
                  Yuwei Jiang and
                  Defei Bu and
                  Li Wei and
                  Haihang Jing and
                  Hongyin Tang and
                  Xin Chen and
                  Xiangzhou Huang and
                  Fengcun Li and
                  Rongxiang Weng and
                  Yulei Qian and
                  Yifan Lu and
                  Yerui Sun and
                  Jingang Wang and
                  Yuchen Xie and
                  Xunliang Cai},
  title        = {Efficient Context Scaling with LongCat ZigZag Attention},
  journal      = {CoRR},
  volume       = {abs/2512.23966},
  year         = {2025},
  url          = {https://doi.org/10.48550/arXiv.2512.23966},
  doi          = {10.48550/ARXIV.2512.23966},
  eprinttype   = {arXiv},
  eprint       = {2512.23966},
  bibsource    = {dblp computer science bibliography, https://dblp.org}
}

@inproceedings{VoitaTMST19,
  author       = {Elena Voita and
                  David Talbot and
                  Fedor Moiseev and
                  Rico Sennrich and
                  Ivan Titov},
  editor       = {Anna Korhonen and
                  David R. Traum and
                  Llu{\'{\i}}s M{\`{a}}rquez},
  title        = {Analyzing Multi-Head Self-Attention: Specialized Heads Do the Heavy
                  Lifting, the Rest Can Be Pruned},
  booktitle    = {Proceedings of the 57th Conference of the Association for Computational
                  Linguistics, {ACL} 2019, Florence, Italy, July 28- August 2, 2019,
                  Volume 1: Long Papers},
  pages        = {5797--5808},
  publisher    = {Association for Computational Linguistics},
  year         = {2019},
  url          = {https://doi.org/10.18653/v1/p19-1580},
  doi          = {10.18653/V1/P19-1580},
  bibsource    = {dblp computer science bibliography, https://dblp.org}
}

@inproceedings{ClarkKLM19,
  author       = {Kevin Clark and
                  Urvashi Khandelwal and
                  Omer Levy and
                  Christopher D. Manning},
  editor       = {Tal Linzen and
                  Grzegorz Chrupala and
                  Yonatan Belinkov and
                  Dieuwke Hupkes},
  title        = {What Does {BERT} Look at? An Analysis of BERT's Attention},
  booktitle    = {Proceedings of the 2019 {ACL} Workshop BlackboxNLP: Analyzing and
                  Interpreting Neural Networks for NLP, BlackboxNLP@ACL 2019, Florence,
                  Italy, August 1, 2019},
  pages        = {276--286},
  publisher    = {Association for Computational Linguistics},
  year         = {2019},
  url          = {https://doi.org/10.18653/v1/W19-4828},
  doi          = {10.18653/V1/W19-4828},
  bibsource    = {dblp computer science bibliography, https://dblp.org}
}

@inproceedings{XiaGZ024,
  author       = {Mengzhou Xia and
                  Tianyu Gao and
                  Zhiyuan Zeng and
                  Danqi Chen},
  title        = {Sheared LLaMA: Accelerating Language Model Pre-training via Structured
                  Pruning},
  booktitle    = {The Twelfth International Conference on Learning Representations,
                  {ICLR} 2024, Vienna, Austria, May 7-11, 2024},
  publisher    = {OpenReview.net},
  year         = {2024},
  url          = {https://openreview.net/forum?id=09iOdaeOzp},
  bibsource    = {dblp computer science bibliography, https://dblp.org}
}

@article{abs-2004-05150,
  author       = {Iz Beltagy and
                  Matthew E. Peters and
                  Arman Cohan},
  title        = {Longformer: The Long-Document Transformer},
  journal      = {CoRR},
  volume       = {abs/2004.05150},
  year         = {2020},
  url          = {https://arxiv.org/abs/2004.05150},
  eprinttype   = {arXiv},
  eprint       = {2004.05150},
  bibsource    = {dblp computer science bibliography, https://dblp.org}
}

@inproceedings{ZaheerGDAAOPRWY20,
  author       = {Manzil Zaheer and
                  Guru Guruganesh and
                  Kumar Avinava Dubey and
                  Joshua Ainslie and
                  Chris Alberti and
                  Santiago Onta{\~{n}}{\'{o}}n and
                  Philip Pham and
                  Anirudh Ravula and
                  Qifan Wang and
                  Li Yang and
                  Amr Ahmed},
  editor       = {Hugo Larochelle and
                  Marc'Aurelio Ranzato and
                  Raia Hadsell and
                  Maria{-}Florina Balcan and
                  Hsuan{-}Tien Lin},
  title        = {Big Bird: Transformers for Longer Sequences},
  booktitle    = {Advances in Neural Information Processing Systems 33: Annual Conference
                  on Neural Information Processing Systems 2020, NeurIPS 2020, December
                  6-12, 2020, virtual},
  year         = {2020},
  url          = {https://proceedings.neurips.cc/paper/2020/hash/c8512d142a2d849725f31a9a7a361ab9-Abstract.html},
  bibsource    = {dblp computer science bibliography, https://dblp.org}
}

@article{abs-1904-10509,
  author       = {Rewon Child and
                  Scott Gray and
                  Alec Radford and
                  Ilya Sutskever},
  title        = {Generating Long Sequences with Sparse Transformers},
  journal      = {CoRR},
  volume       = {abs/1904.10509},
  year         = {2019},
  url          = {http://arxiv.org/abs/1904.10509},
  eprinttype   = {arXiv},
  eprint       = {1904.10509},
  bibsource    = {dblp computer science bibliography, https://dblp.org}
}

@inproceedings{KatharopoulosV020,
  author       = {Angelos Katharopoulos and
                  Apoorv Vyas and
                  Nikolaos Pappas and
                  Fran{\c{c}}ois Fleuret},
  title        = {Transformers are RNNs: Fast Autoregressive Transformers with Linear
                  Attention},
  booktitle    = {Proceedings of the 37th International Conference on Machine Learning,
                  {ICML} 2020, 13-18 July 2020, Virtual Event},
  series       = {Proceedings of Machine Learning Research},
  pages        = {5156--5165},
  publisher    = {{PMLR}},
  year         = {2020},
  url          = {http://proceedings.mlr.press/v119/katharopoulos20a.html},
  bibsource    = {dblp computer science bibliography, https://dblp.org}
}

@inproceedings{KitaevKL20,
  author       = {Nikita Kitaev and
                  Lukasz Kaiser and
                  Anselm Levskaya},
  title        = {Reformer: The Efficient Transformer},
  booktitle    = {8th International Conference on Learning Representations, {ICLR} 2020,
                  Addis Ababa, Ethiopia, April 26-30, 2020},
  publisher    = {OpenReview.net},
  year         = {2020},
  url          = {https://openreview.net/forum?id=rkgNKkHtvB},
  bibsource    = {dblp computer science bibliography, https://dblp.org}
}

@article{RoySVG21,
  author       = {Aurko Roy and
                  Mohammad Saffar and
                  Ashish Vaswani and
                  David Grangier},
  title        = {Efficient Content-Based Sparse Attention with Routing Transformers},
  journal      = {Trans. Assoc. Comput. Linguistics},
  volume       = {9},
  pages        = {53--68},
  year         = {2021},
  url          = {https://doi.org/10.1162/tacl\_a\_00353},
  doi          = {10.1162/TACL\_A\_00353},
  bibsource    = {dblp computer science bibliography, https://dblp.org}
}

@article{abs-2312-00752,
  author       = {Albert Gu and
                  Tri Dao},
  title        = {Mamba: Linear-Time Sequence Modeling with Selective State Spaces},
  journal      = {CoRR},
  volume       = {abs/2312.00752},
  year         = {2023},
  url          = {https://doi.org/10.48550/arXiv.2312.00752},
  doi          = {10.48550/ARXIV.2312.00752},
  eprinttype   = {arXiv},
  eprint       = {2312.00752},
  bibsource    = {dblp computer science bibliography, https://dblp.org}
}

@inproceedings{LenzLABMPAAFPGJ25,
  author       = {Barak Lenz and
                  Opher Lieber and
                  Alan Arazi and
                  Amir Bergman and
                  Avshalom Manevich and
                  Barak Peleg and
                  Ben Aviram and
                  Chen Almagor and
                  Clara Fridman and
                  Dan Padnos and
                  Daniel Gissin and
                  Daniel Jannai and
                  Dor Muhlgay and
                  Dor Zimberg and
                  Edden M. Gerber and
                  Elad Dolev and
                  Eran Krakovsky and
                  Erez Safahi and
                  Erez Schwartz and
                  Gal Cohen and
                  et al.},
  title        = {Jamba: Hybrid Transformer-Mamba Language Models},
  booktitle    = {The Thirteenth International Conference on Learning Representations,
                  {ICLR} 2025, Singapore, April 24-28, 2025},
  publisher    = {OpenReview.net},
  year         = {2025},
  url          = {https://openreview.net/forum?id=JFPaD7lpBD},
  bibsource    = {dblp computer science bibliography, https://dblp.org}
}

@inproceedings{XiaoTZGYTF025,
  author       = {Guangxuan Xiao and
                  Jiaming Tang and
                  Jingwei Zuo and
                  Junxian Guo and
                  Shang Yang and
                  Haotian Tang and
                  Yao Fu and
                  Song Han},
  title        = {DuoAttention: Efficient Long-Context {LLM} Inference with Retrieval
                  and Streaming Heads},
  booktitle    = {The Thirteenth International Conference on Learning Representations,
                  {ICLR} 2025, Singapore, April 24-28, 2025},
  publisher    = {OpenReview.net},
  year         = {2025},
  url          = {https://openreview.net/forum?id=cFu7ze7xUm},
  bibsource    = {dblp computer science bibliography, https://dblp.org}
}

@article{MiMo-V2,
  author       = {LLM{-}Core Xiaomi},
  title        = {MiMo-V2-Flash Technical Report},
  journal      = {CoRR},
  volume       = {abs/2601.02780},
  year         = {2026},
  url          = {https://doi.org/10.48550/arXiv.2601.02780},
  doi          = {10.48550/ARXIV.2601.02780},
  eprinttype   = {arXiv},
  eprint       = {2601.02780},
  bibsource    = {dblp computer science bibliography, https://dblp.org}
}

@inproceedings{piqa,
  author       = {Yonatan Bisk and
                  Rowan Zellers and
                  Ronan Le Bras and
                  Jianfeng Gao and
                  Yejin Choi},
  title        = {{PIQA:} Reasoning about Physical Commonsense in Natural Language},
  booktitle    = {The Thirty-Fourth {AAAI} Conference on Artificial Intelligence, {AAAI}
                  2020, The Thirty-Second Innovative Applications of Artificial Intelligence
                  Conference, {IAAI} 2020, The Tenth {AAAI} Symposium on Educational
                  Advances in Artificial Intelligence, {EAAI} 2020, New York, NY, USA,
                  February 7-12, 2020},
  pages        = {7432--7439},
  publisher    = {{AAAI} Press},
  year         = {2020},
  url          = {https://doi.org/10.1609/aaai.v34i05.6239},
  doi          = {10.1609/AAAI.V34I05.6239},
  bibsource    = {dblp computer science bibliography, https://dblp.org}
}

@inproceedings{MMLU,
  author       = {Dan Hendrycks and
                  Collin Burns and
                  Steven Basart and
                  Andy Zou and
                  Mantas Mazeika and
                  Dawn Song and
                  Jacob Steinhardt},
  title        = {Measuring Massive Multitask Language Understanding},
  booktitle    = {9th International Conference on Learning Representations, {ICLR} 2021,
                  Virtual Event, Austria, May 3-7, 2021},
  publisher    = {OpenReview.net},
  year         = {2021},
  url          = {https://openreview.net/forum?id=d7KBjmI3GmQ},
  bibsource    = {dblp computer science bibliography, https://dblp.org}
}

@inproceedings{logiqa,
  author       = {Jian Liu and
                  Leyang Cui and
                  Hanmeng Liu and
                  Dandan Huang and
                  Yile Wang and
                  Yue Zhang},
  editor       = {Christian Bessiere},
  title        = {LogiQA: {A} Challenge Dataset for Machine Reading Comprehension with
                  Logical Reasoning},
  booktitle    = {Proceedings of the Twenty-Ninth International Joint Conference on
                  Artificial Intelligence, {IJCAI} 2020},
  pages        = {3622--3628},
  publisher    = {ijcai.org},
  year         = {2020},
  url          = {https://doi.org/10.24963/ijcai.2020/501},
  doi          = {10.24963/IJCAI.2020/501},
  bibsource    = {dblp computer science bibliography, https://dblp.org}
}

@inproceedings{CSQA,
  author       = {Alon Talmor and
                  Jonathan Herzig and
                  Nicholas Lourie and
                  Jonathan Berant},
  editor       = {Jill Burstein and
                  Christy Doran and
                  Thamar Solorio},
  title        = {CommonsenseQA: {A} Question Answering Challenge Targeting Commonsense
                  Knowledge},
  booktitle    = {Proceedings of the 2019 Conference of the North American Chapter of
                  the Association for Computational Linguistics: Human Language Technologies,
                  {NAACL-HLT} 2019, Minneapolis, MN, USA, June 2-7, 2019, Volume 1 (Long
                  and Short Papers)},
  pages        = {4149--4158},
  publisher    = {Association for Computational Linguistics},
  year         = {2019},
  url          = {https://doi.org/10.18653/v1/n19-1421},
  doi          = {10.18653/V1/N19-1421},
  bibsource    = {dblp computer science bibliography, https://dblp.org}
}

@inproceedings{siqa,
  author       = {Maarten Sap and
                  Hannah Rashkin and
                  Derek Chen and
                  Ronan Le Bras and
                  Yejin Choi},
  editor       = {Kentaro Inui and
                  Jing Jiang and
                  Vincent Ng and
                  Xiaojun Wan},
  title        = {Social IQa: Commonsense Reasoning about Social Interactions},
  booktitle    = {Proceedings of the 2019 Conference on Empirical Methods in Natural
                  Language Processing and the 9th International Joint Conference on
                  Natural Language Processing, {EMNLP-IJCNLP} 2019, Hong Kong, China,
                  November 3-7, 2019},
  pages        = {4462--4472},
  publisher    = {Association for Computational Linguistics},
  year         = {2019},
  url          = {https://doi.org/10.18653/v1/D19-1454},
  doi          = {10.18653/V1/D19-1454},
  bibsource    = {dblp computer science bibliography, https://dblp.org}
}

@article{arc,
  author       = {Peter Clark and
                  Isaac Cowhey and
                  Oren Etzioni and
                  Tushar Khot and
                  Ashish Sabharwal and
                  Carissa Schoenick and
                  Oyvind Tafjord},
  title        = {Think you have Solved Question Answering? Try ARC, the {AI2} Reasoning
                  Challenge},
  journal      = {CoRR},
  volume       = {abs/1803.05457},
  year         = {2018},
  url          = {http://arxiv.org/abs/1803.05457},
  eprinttype   = {arXiv},
  eprint       = {1803.05457},
  bibsource    = {dblp computer science bibliography, https://dblp.org}
}

@inproceedings{hella,
  author       = {Rowan Zellers and
                  Ari Holtzman and
                  Yonatan Bisk and
                  Ali Farhadi and
                  Yejin Choi},
  editor       = {Anna Korhonen and
                  David R. Traum and
                  Llu{\'{\i}}s M{\`{a}}rquez},
  title        = {HellaSwag: Can a Machine Really Finish Your Sentence?},
  booktitle    = {Proceedings of the 57th Conference of the Association for Computational
                  Linguistics, {ACL} 2019, Florence, Italy, July 28- August 2, 2019,
                  Volume 1: Long Papers},
  pages        = {4791--4800},
  publisher    = {Association for Computational Linguistics},
  year         = {2019},
  url          = {https://doi.org/10.18653/v1/p19-1472},
  doi          = {10.18653/V1/P19-1472},
  bibsource    = {dblp computer science bibliography, https://dblp.org}
}

@misc{webqa-cn,
      title={Dataset and Neural Recurrent Sequence Labeling Model for Open-Domain Factoid Question Answering}, 
      author={Peng Li and Wei Li and Zhengyan He and Xuguang Wang and Ying Cao and Jie Zhou and Wei Xu},
      year={2016},
      eprint={1607.06275},
      archivePrefix={arXiv},
      primaryClass={cs.CL},
      url={https://arxiv.org/abs/1607.06275}, 
}

@inproceedings{CSL,
  author       = {Yudong Li and
                  Yuqing Zhang and
                  Zhe Zhao and
                  Linlin Shen and
                  Weijie Liu and
                  Weiquan Mao and
                  Hui Zhang},
  editor       = {Nicoletta Calzolari and
                  Chu{-}Ren Huang and
                  Hansaem Kim and
                  James Pustejovsky and
                  Leo Wanner and
                  Key{-}Sun Choi and
                  Pum{-}Mo Ryu and
                  Hsin{-}Hsi Chen and
                  Lucia Donatelli and
                  Heng Ji and
                  Sadao Kurohashi and
                  Patrizia Paggio and
                  Nianwen Xue and
                  Seokhwan Kim and
                  Younggyun Hahm and
                  Zhong He and
                  Tony Kyungil Lee and
                  Enrico Santus and
                  Francis Bond and
                  Seung{-}Hoon Na},
  title        = {{CSL:} {A} Large-scale Chinese Scientific Literature Dataset},
  booktitle    = {Proceedings of the 29th International Conference on Computational
                  Linguistics, {COLING} 2022, Gyeongju, Republic of Korea, October 12-17,
                  2022},
  pages        = {3917--3923},
  publisher    = {International Committee on Computational Linguistics},
  year         = {2022},
  url          = {https://aclanthology.org/2022.coling-1.344},
  bibsource    = {dblp computer science bibliography, https://dblp.org}
}

@article{LOT,
  author       = {Jian Guan and
                  Zhuoer Feng and
                  Yamei Chen and
                  Ruilin He and
                  Xiaoxi Mao and
                  Changjie Fan and
                  Minlie Huang},
  title        = {{LOT:} {A} Story-Centric Benchmark for Evaluating Chinese Long Text
                  Understanding and Generation},
  journal      = {Trans. Assoc. Comput. Linguistics},
  volume       = {10},
  pages        = {434--451},
  year         = {2022},
  url          = {https://doi.org/10.1162/tacl\_a\_00469},
  doi          = {10.1162/TACL\_A\_00469},
  bibsource    = {dblp computer science bibliography, https://dblp.org}
}

@inproceedings{louizos2018learning,
  title={Learning Sparse Neural Networks through L\_0 Regularization},
  author={Louizos, Christos and Welling, Max and Kingma, Diederik P},
  booktitle={International Conference on Learning Representations},
  year={2018}
}

@article{Switch_Attention,
  author       = {Yusheng Zhao and
                  Hourun Li and
                  Bohan Wu and
                  Jingyang Yuan and
                  Meng Zhang and
                  Yichun Yin and
                  Lifeng Shang and
                  Ming Zhang},
  title        = {Switch Attention: Towards Dynamic and Fine-grained Hybrid Transformers},
  journal      = {CoRR},
  volume       = {abs/2603.26380},
  year         = {2026},
  url          = {https://doi.org/10.48550/arXiv.2603.26380},
  doi          = {10.48550/ARXIV.2603.26380},
  eprinttype   = {arXiv},
  eprint       = {2603.26380},
  bibsource    = {dblp computer science bibliography, https://dblp.org}
}

@inproceedings{chen-etal-2025-improving-reasoning,
    title = "Improving Reasoning Capabilities in Small Models through Mixture-of-layers Distillation with Stepwise Attention on Key Information",
    author = "Chen, Yao  and
      Sheng, Jiawei  and
      Zhang, Wenyuan  and
      Liu, Tingwen",
    editor = "Christodoulopoulos, Christos  and
      Chakraborty, Tanmoy  and
      Rose, Carolyn  and
      Peng, Violet",
    booktitle = "Proceedings of the 2025 Conference on Empirical Methods in Natural Language Processing",
    month = nov,
    year = "2025",
    address = "Suzhou, China",
    publisher = "Association for Computational Linguistics",
    url = "https://aclanthology.org/2025.emnlp-main.250/",
    doi = "10.18653/v1/2025.emnlp-main.250",
    pages = "4952--4971",
    ISBN = "979-8-89176-332-6"
}

@inproceedings{WeberFAOAALNYAA24,
  author       = {Maurice Weber and
                  Daniel Y. Fu and
                  Quentin Anthony and
                  Yonatan Oren and
                  Shane Adams and
                  Anton Alexandrov and
                  Xiaozhong Lyu and
                  Huu Nguyen and
                  Xiaozhe Yao and
                  Virginia Adams and
                  Ben Athiwaratkun and
                  Rahul Chalamala and
                  Kezhen Chen and
                  Max Ryabinin and
                  Tri Dao and
                  Percy Liang and
                  Christopher R{\'{e}} and
                  Irina Rish and
                  Ce Zhang},
  editor       = {Amir Globersons and
                  Lester Mackey and
                  Danielle Belgrave and
                  Angela Fan and
                  Ulrich Paquet and
                  Jakub M. Tomczak and
                  Cheng Zhang},
  title        = {RedPajama: an Open Dataset for Training Large Language Models},
  booktitle    = {Advances in Neural Information Processing Systems 38: Annual Conference
                  on Neural Information Processing Systems 2024, NeurIPS 2024, Vancouver,
                  BC, Canada, December 10 - 15, 2024},
  year         = {2024},
  url          = {http://papers.nips.cc/paper\_files/paper/2024/hash/d34497330b1fd6530f7afd86d0df9f76-Abstract-Datasets\_and\_Benchmarks\_Track.html},
  bibsource    = {dblp computer science bibliography, https://dblp.org}
}

@article{abs-2501-08197,
  author       = {Yijiong Yu and
                  Ziyun Dai and
                  Zekun Wang and
                  Wei Wang and
                  Ran Chen and
                  Ji Pei},
  title        = {OpenCSG Chinese Corpus: {A} Series of High-quality Chinese Datasets
                  for {LLM} Training},
  journal      = {CoRR},
  volume       = {abs/2501.08197},
  year         = {2025},
  url          = {https://doi.org/10.48550/arXiv.2501.08197},
  doi          = {10.48550/ARXIV.2501.08197},
  eprinttype   = {arXiv},
  eprint       = {2501.08197},
  bibsource    = {dblp computer science bibliography, https://dblp.org}
}

@inproceedings{BaiLZL0HDLZHDTL24,
  author       = {Yushi Bai and
                  Xin Lv and
                  Jiajie Zhang and
                  Hongchang Lyu and
                  Jiankai Tang and
                  Zhidian Huang and
                  Zhengxiao Du and
                  Xiao Liu and
                  Aohan Zeng and
                  Lei Hou and
                  Yuxiao Dong and
                  Jie Tang and
                  Juanzi Li},
  editor       = {Lun{-}Wei Ku and
                  Andre Martins and
                  Vivek Srikumar},
  title        = {LongBench: {A} Bilingual, Multitask Benchmark for Long Context Understanding},
  booktitle    = {Proceedings of the 62nd Annual Meeting of the Association for Computational
                  Linguistics (Volume 1: Long Papers), {ACL} 2024, Bangkok, Thailand,
                  August 11-16, 2024},
  pages        = {3119--3137},
  publisher    = {Association for Computational Linguistics},
  year         = {2024},
  url          = {https://doi.org/10.18653/v1/2024.acl-long.172},
  doi          = {10.18653/V1/2024.ACL-LONG.172},
  bibsource    = {dblp computer science bibliography, https://dblp.org}
}

@article{KociskySBDHMG18,
  author       = {Tom{\'{a}}s Kocisk{\'{y}} and
                  Jonathan Schwarz and
                  Phil Blunsom and
                  Chris Dyer and
                  Karl Moritz Hermann and
                  G{\'{a}}bor Melis and
                  Edward Grefenstette},
  title        = {The NarrativeQA Reading Comprehension Challenge},
  journal      = {Trans. Assoc. Comput. Linguistics},
  volume       = {6},
  pages        = {317--328},
  year         = {2018},
  url          = {https://doi.org/10.1162/tacl\_a\_00023},
  doi          = {10.1162/TACL\_A\_00023},
  bibsource    = {dblp computer science bibliography, https://dblp.org}
}

@inproceedings{HuangCPJW21,
  author       = {Luyang Huang and
                  Shuyang Cao and
                  Nikolaus Nova Parulian and
                  Heng Ji and
                  Lu Wang},
  editor       = {Kristina Toutanova and
                  Anna Rumshisky and
                  Luke Zettlemoyer and
                  Dilek Hakkani{-}T{\"{u}}r and
                  Iz Beltagy and
                  Steven Bethard and
                  Ryan Cotterell and
                  Tanmoy Chakraborty and
                  Yichao Zhou},
  title        = {Efficient Attentions for Long Document Summarization},
  booktitle    = {Proceedings of the 2021 Conference of the North American Chapter of
                  the Association for Computational Linguistics: Human Language Technologies,
                  {NAACL-HLT} 2021, Online, June 6-11, 2021},
  pages        = {1419--1436},
  publisher    = {Association for Computational Linguistics},
  year         = {2021},
  url          = {https://doi.org/10.18653/v1/2021.naacl-main.112},
  doi          = {10.18653/V1/2021.NAACL-MAIN.112},
  bibsource    = {dblp computer science bibliography, https://dblp.org}
}

@inproceedings{Yang0ZBCSM18,
  author       = {Zhilin Yang and
                  Peng Qi and
                  Saizheng Zhang and
                  Yoshua Bengio and
                  William W. Cohen and
                  Ruslan Salakhutdinov and
                  Christopher D. Manning},
  editor       = {Ellen Riloff and
                  David Chiang and
                  Julia Hockenmaier and
                  Jun'ichi Tsujii},
  title        = {HotpotQA: {A} Dataset for Diverse, Explainable Multi-hop Question
                  Answering},
  booktitle    = {Proceedings of the 2018 Conference on Empirical Methods in Natural
                  Language Processing, Brussels, Belgium, October 31 - November 4, 2018},
  pages        = {2369--2380},
  publisher    = {Association for Computational Linguistics},
  year         = {2018},
  url          = {https://doi.org/10.18653/v1/d18-1259},
  doi          = {10.18653/V1/D18-1259},
  bibsource    = {dblp computer science bibliography, https://dblp.org}
}

@article{LiuLHPBPL24,
  author       = {Nelson F. Liu and
                  Kevin Lin and
                  John Hewitt and
                  Ashwin Paranjape and
                  Michele Bevilacqua and
                  Fabio Petroni and
                  Percy Liang},
  title        = {Lost in the Middle: How Language Models Use Long Contexts},
  journal      = {Trans. Assoc. Comput. Linguistics},
  volume       = {12},
  pages        = {157--173},
  year         = {2024},
  url          = {https://doi.org/10.1162/tacl\_a\_00638},
  doi          = {10.1162/TACL\_A\_00638},
  bibsource    = {dblp computer science bibliography, https://dblp.org}
}

@inproceedings{KwonLZ0ZY0ZS23,
  author       = {Woosuk Kwon and
                  Zhuohan Li and
                  Siyuan Zhuang and
                  Ying Sheng and
                  Lianmin Zheng and
                  Cody Hao Yu and
                  Joseph Gonzalez and
                  Hao Zhang and
                  Ion Stoica},
  editor       = {Jason Flinn and
                  Margo I. Seltzer and
                  Peter Druschel and
                  Antoine Kaufmann and
                  Jonathan Mace},
  title        = {Efficient Memory Management for Large Language Model Serving with
                  PagedAttention},
  booktitle    = {Proceedings of the 29th Symposium on Operating Systems Principles,
                  {SOSP} 2023, Koblenz, Germany, October 23-26, 2023},
  pages        = {611--626},
  publisher    = {{ACM}},
  year         = {2023},
  url          = {https://doi.org/10.1145/3600006.3613165},
  doi          = {10.1145/3600006.3613165},
  bibsource    = {dblp computer science bibliography, https://dblp.org}
}

@inproceedings{XiaoZ0XLZ0024,
  author       = {Chaojun Xiao and
                  Pengle Zhang and
                  Xu Han and
                  Guangxuan Xiao and
                  Yankai Lin and
                  Zhengyan Zhang and
                  Zhiyuan Liu and
                  Maosong Sun},
  editor       = {Amir Globersons and
                  Lester Mackey and
                  Danielle Belgrave and
                  Angela Fan and
                  Ulrich Paquet and
                  Jakub M. Tomczak and
                  Cheng Zhang},
  title        = {InfLLM: Training-Free Long-Context Extrapolation for LLMs with an
                  Efficient Context Memory},
  booktitle    = {Advances in Neural Information Processing Systems 38: Annual Conference
                  on Neural Information Processing Systems 2024, NeurIPS 2024, Vancouver,
                  BC, Canada, December 10 - 15, 2024},
  year         = {2024},
  url          = {http://papers.nips.cc/paper\_files/paper/2024/hash/d842425e4bf79ba039352da0f658a906-Abstract-Conference.html},
  bibsource    = {dblp computer science bibliography, https://dblp.org}
}

@article{abs-2603-26380,
  author       = {Yusheng Zhao and
                  Hourun Li and
                  Bohan Wu and
                  Jingyang Yuan and
                  Meng Zhang and
                  Yichun Yin and
                  Lifeng Shang and
                  Ming Zhang},
  title        = {Switch Attention: Towards Dynamic and Fine-grained Hybrid Transformers},
  journal      = {CoRR},
  volume       = {abs/2603.26380},
  year         = {2026},
  url          = {https://doi.org/10.48550/arXiv.2603.26380},
  doi          = {10.48550/ARXIV.2603.26380},
  eprinttype   = {arXiv},
  eprint       = {2603.26380},
  bibsource    = {dblp computer science bibliography, https://dblp.org}
}

% \newpage
\clearpage
\appendix

\section{Comparison with Existing Approaches}
\label{app.comparison}

Table~\ref{tab:comparison_settings} compares \aname with rule-based interleaving~\cite{abs-2310-06825,abs-2408-00118} and LoZA~\cite{abs-2512-23966} along several design dimensions.
Rule-based methods and LoZA both allocate at the layer level, so all KV heads within a layer share the same attention type; \aname supports both layer-wise and KV-head-wise granularity under the same framework, and the head-wise variant delivers most of the empirical gain.
On sparsity control, LoZA selects layers by ranking scalar weights post-hoc and reports only a single $50\%$ ratio; \aname instead enforces $\hat{\rho}(z) = \rho$ as a Lagrangian equality constraint, allowing the user to specify any $\rho$ in advance, with convergence verified in Appendix~\ref{appdx:lagrangian-convergence}.
On training integration, LoZA freezes model weights during calibration and then mid-trains under the frozen pattern; \aname jointly optimizes $\theta$ and $\alpha$ during Stage~1 and transitions into continued pre-training under the binarized masks.

Two additional methods are worth noting.
SwiAttn~\cite{Switch_Attention} dynamically routes each token to FA or SWA via per-layer routers, but must retain a unified KV cache because any token may require full attention, reducing compute but not memory; \aname's fixed binarized masks ensure that SWA-allocated heads retain only window-sized KV, yielding genuine KV cache reduction.
DuoAttention~\cite{XiaoTZGYTF025} operates at head granularity and achieves similar memory savings, but relies on synthetic long-range retrieval data and output-deviation minimization with frozen model weights; the realized sparsity is controlled indirectly via L1 regularization and thresholding rather than an explicit target.
\aname aims to find effective hybrid configurations during the pre-training stage using only standard pre-training data, jointly adapting both the allocation masks and model weights in a single optimization framework.

\section{Training Details}
\label{app.Training Details}

\paragraph{Hyperparameters.}
We follow the standard configuration of~\citet{louizos2018learning} and set $\beta = 2/3$, $\zeta = 1.1$, $\gamma = -0.1$ across all experiments without further tuning.
The Lagrange multipliers $\lambda$ and $\phi$ are both initialized to zero and updated by gradient ascent jointly with the model and mask parameters. 
For Stage~1 mask learning, we use the Adam optimizer ($\beta_1 = 0.9$, $\beta_2 = 0.95$, $\epsilon = 10^{-8}$) with a cosine learning rate schedule, a peak learning rate of $3 \times 10^{-4}$, a minimum learning rate of $3 \times 10^{-5}$, and 300 warmup steps.
The global batch size is 128 with a maximum sequence length of 8,192. Training runs for 1,000 steps, corresponding to approximately 1B tokens. 
For Stage~2 continued pre-training, we use the same optimizer with a WSD (Warmup-Stable-Decay) learning rate schedule, a peak learning rate of $3 \times 10^{-4}$, and 2,000 warmup steps.
The global batch size is increased to 512 with the same sequence length of 8,192. All experiments use mixed-precision training in BF16 with a gradient clipping threshold of 1.0.

\begin{table}[t]
\centering
\small
\resizebox{\columnwidth}{!}{
\begin{tabular}{lccccccc}
\toprule
\textbf{Size} & \textbf{L} & \textbf{Hid.} & \textbf{FFN} & \textbf{Q} & \textbf{KV} & \textbf{Dim.} & \textbf{Vocab} \\
\midrule
0.6B & 28 & 1024 & 3072 & 16 & 8 & 64  & 102400 \\
1.7B & 28 & 2048 & 6144 & 16 & 8 & 128 & 102400 \\
\bottomrule
\end{tabular}
}
\caption{Model architecture configurations. L denotes the number of layers; Hid. denotes the hidden size; FFN denotes the intermediate feed-forward dimension; Q/KV denote query and KV head counts; Dim. denotes the per-head dimension.}
\label{tab:model_para}
\end{table}

\begin{figure}[t]
    \centering
    \includegraphics[width=\columnwidth]{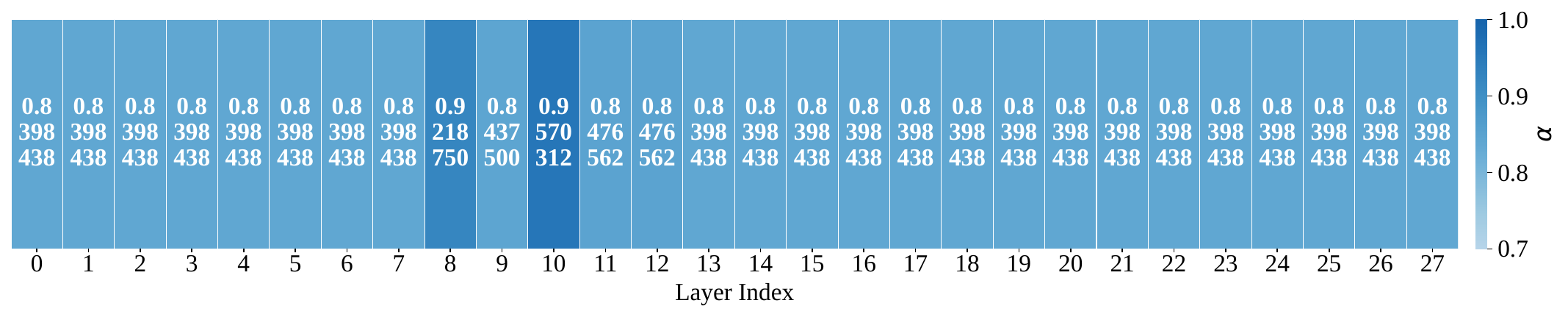}
    \caption{Learned scalar gates $\alpha_i$ of the ablation variant across 28 layers. Nearly all layers converge to the same value, showing minimal differentiation.}
    \label{fig:ablation_alpha}
\end{figure}

\begin{figure*}[t]
    \centering
    \begin{subfigure}[b]{0.32\textwidth}
        \centering
        \includegraphics[width=\textwidth]{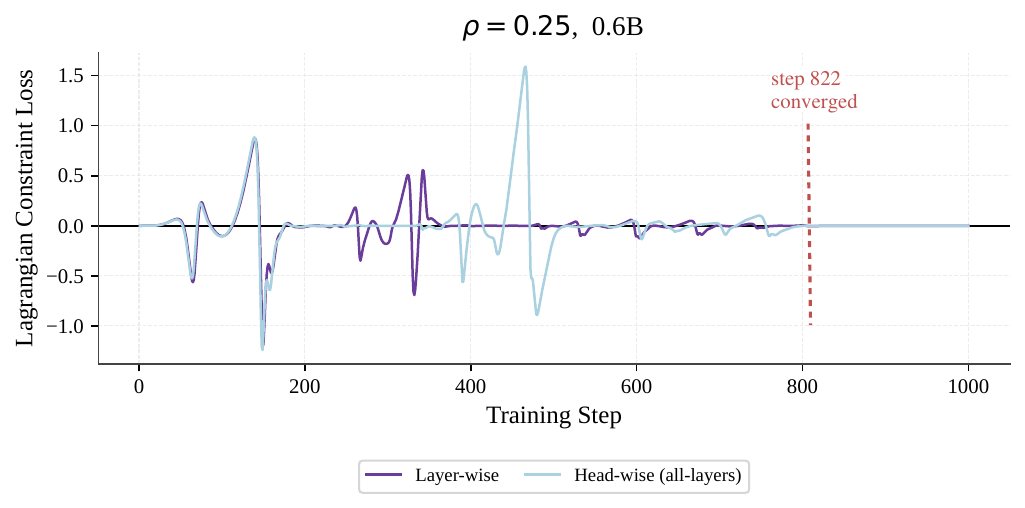}
        \caption{$\rho = 0.25$}
        \label{fig:lagrangian_06b_025}
    \end{subfigure}
    \hfill
    \begin{subfigure}[b]{0.32\textwidth}
        \centering
        \includegraphics[width=\textwidth]{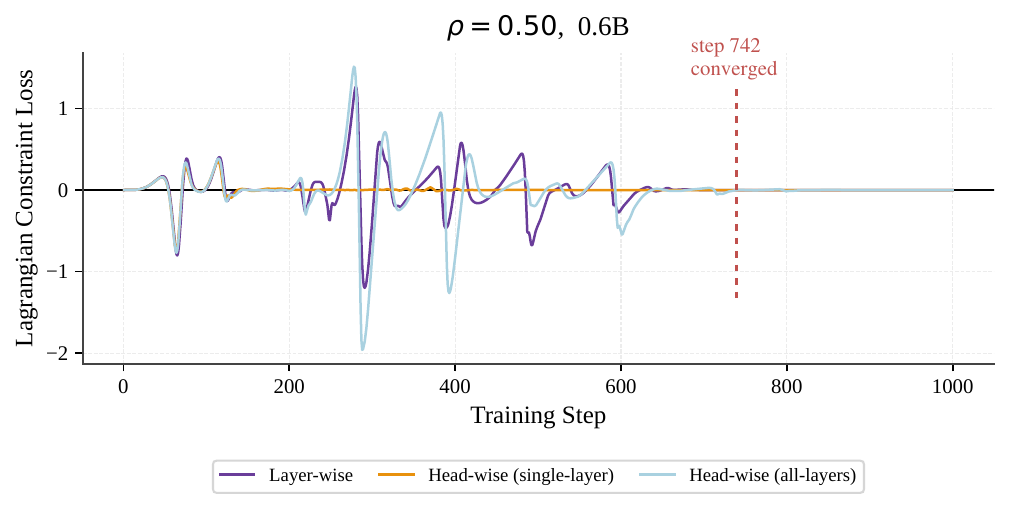}
        \caption{$\rho = 0.50$}
        \label{fig:lagrangian_06b_050}
    \end{subfigure}
    \hfill
    \begin{subfigure}[b]{0.32\textwidth}
        \centering
        \includegraphics[width=\textwidth]{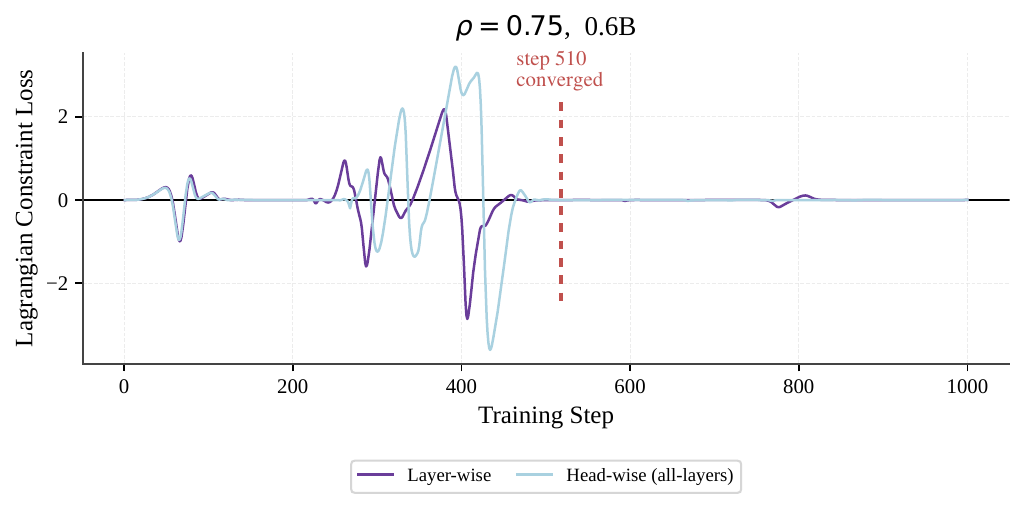}
        \caption{$\rho = 0.75$}
        \label{fig:lagrangian_06b_075}
    \end{subfigure}
    \caption{Lagrangian constraint loss on 0.6B at $\rho \in \{0.25, 0.50, 0.75\}$. 
    The $\rho = 0.25$ and $\rho = 0.75$ settings use layer-wise and head-wise (all-layers) allocation; $\rho = 0.50$ additionally includes head-wise (single-layer). Dashed vertical lines mark the approximate convergence step for the slowest configuration in each panel.}
    \label{fig:lagrangian_06b_all}
\end{figure*}

\begin{figure}[t]
    \centering
    \begin{subfigure}[b]{\columnwidth}
        \centering
        \includegraphics[width=\textwidth]{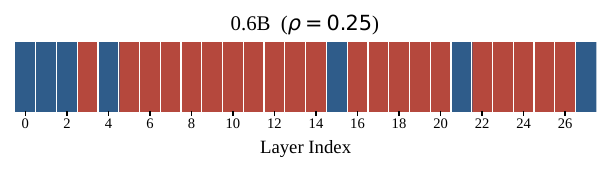}
    \end{subfigure}

     \begin{subfigure}[b]{\columnwidth}
        \centering
        \includegraphics[width=\textwidth]{Figures/analysis_of_pattern/exp6p1_layer_swa_distribution.pdf}
    \end{subfigure}
    
    \begin{subfigure}[b]{\columnwidth}
        \centering
        \includegraphics[width=\textwidth]{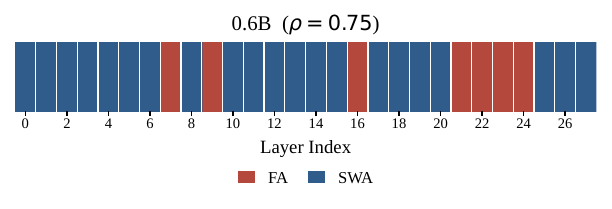}
    \end{subfigure}
    \caption{
    Learned layer-wise FA/SWA allocation on 0.6B at $\rho \in \{0.25, 0.50, 0.75\}$.
     Each cell indicates whether a layer uses \textcolor{customred}{FA (red)} or \textcolor{customblue}{SWA (blue)}.
    }
    \label{fig:layerwise_025_075}
\end{figure}

\begin{table*}[ht]
    \centering
    \renewcommand{\arraystretch}{1.15}
    \resizebox{0.95\textwidth}{!}{%
    \begin{tabular}{lccc}
    \toprule
     & Rule-based~\cite{abs-2310-06825,abs-2408-00118} & LoZA~\cite{abs-2512-23966} & \aname \\
    \midrule
    Allocation method & Hand-crafted & Scalar calibration & L0 + Lagrangian \\
    Granularity & Layer & Layer & Layer \& KV-head \\
    Sparsity control & Fixed by design & Tunable via top-$k$, only 50\% reported & Arbitrary target $\rho$ \\
    Optimization & N/A & Post-hoc scoring & End-to-end differentiable \\
    Pattern analysis & N/A & None & Cross-scale visualization \\
    Training integration & Pre-defined & Post-hoc & Joint with pre-training \\
    \bottomrule
    \end{tabular}%
    }
    \caption{Comparison of \aname with existing methods for hybrid attention allocation.}
    \label{tab:comparison_settings}
\end{table*}

\paragraph{Fair Comparison.}
\aname uses 1B tokens for mask learning in Stage~1, followed by 100B tokens for continued pre-training in Stage~2, for a total of 101B tokens beyond the initial pre-trained checkpoint.
To ensure that the performance gains of \aname are not attributed to this additional training budget, all baselines are trained from the same pre-trained checkpoint for the same 101B tokens across two stages, using the same data mixture, optimizer, learning rate schedule, and batch size as \aname in each corresponding stage.
The training data is drawn from two open-source corpora, RedPajama~\cite{WeberFAOAALNYAA24} and Chinese FineWeb~\cite{abs-2501-08197}.
All reported results are averaged over multiple evaluation runs, so that every method is compared under identical training and evaluation conditions.

\section{Ablation Study: Setup and Analysis}
\label{appdx:ablation-detail}

\paragraph{Ablation Variant Design.}
The ablation variant (w/o L0-Lagrangian) follows the calibration-based paradigm used by LoZA~\cite{abs-2512-23966} and DuoAttention~\cite{XiaoTZGYTF025}.
Since LoZA does not specify its calibration objective in detail and the full DuoAttention setup involves a distillation loss against a dense teacher together with synthetic retrieval data, we adopt a simplified variant for a controlled comparison.
Specifically, our ablation uses the standard language modeling loss and does not rely on synthetic data, thereby isolating the effect of removing the L0-Lagrangian formulation while keeping all other factors aligned with \aname.

\paragraph{Training Configuration.}
For all three sparsity configurations, the 0.6B model starts from the same checkpoint pre-trained from scratch on 40B tokens.
Both \aname and the ablation variant train masks on 1B tokens, followed by 20B tokens of continued pre-training under the resulting fixed attention configuration.
The 20B-token continued pre-training stage consists of 10{,}500 training steps.
The loss trajectories in Figure~\ref{fig:ablation_loss} are displayed starting from step 2,500 to focus on the post-warmup training dynamics.

\paragraph{ConSA vs.\ Dense FA.}
Comparing \aname with Dense FA across the three sparsity levels, the relative position of the two loss curves shifts progressively as $\rho$ increases: \aname trains consistently below Dense FA at $\rho = 0.25$, the two trajectories nearly overlap throughout training at $\rho = 0.50$, and \aname settles slightly above Dense FA at $\rho = 0.75$ where three quarters of the attention units operate with local attention.
This progression shows that the L0-Lagrangian formulation effectively preserves model quality at moderate sparsity levels.

\paragraph{Scalar Gates Fail to Differentiate Attention Units.}
Figure~\ref{fig:ablation_alpha} visualizes the learned scalar gates $\alpha_i$ of the ablation variant across all 28 layers.
Unlike the binary masks produced by \aname, the calibration-based variant yields gate values that are nearly uniform across layers, with most values clustered in a narrow range.
This lack of differentiation suggests that learning a single scalar per attention unit does not provide sufficient signal to distinguish layers that benefit from full attention from those that can operate with local attention.
We note that DuoAttention achieves more differentiated gate values but relies on a considerably more complex setup involving synthetic retrieval data and a multi-component distillation loss relative to a dense teacher, which introduces additional design choices and computational overhead beyond the allocation mechanism itself.

\begin{figure*}[ht]
    \centering
    \begin{subfigure}[b]{0.48\textwidth}
        \centering
        \includegraphics[width=\textwidth]{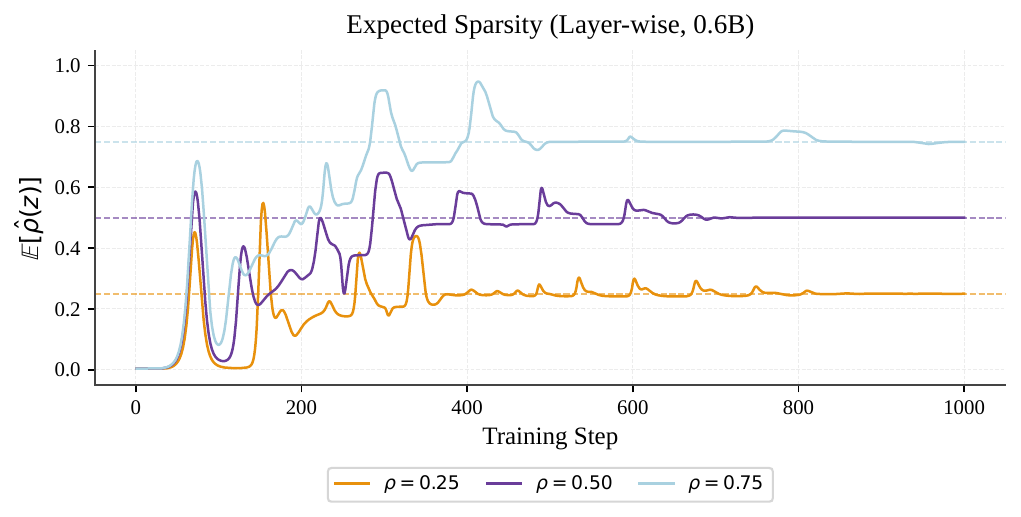}
        \caption{Layer-wise}
        \label{fig:sparsity_layerwise}
    \end{subfigure}
    \hfill
    \begin{subfigure}[b]{0.48\textwidth}
        \centering
        \includegraphics[width=\textwidth]{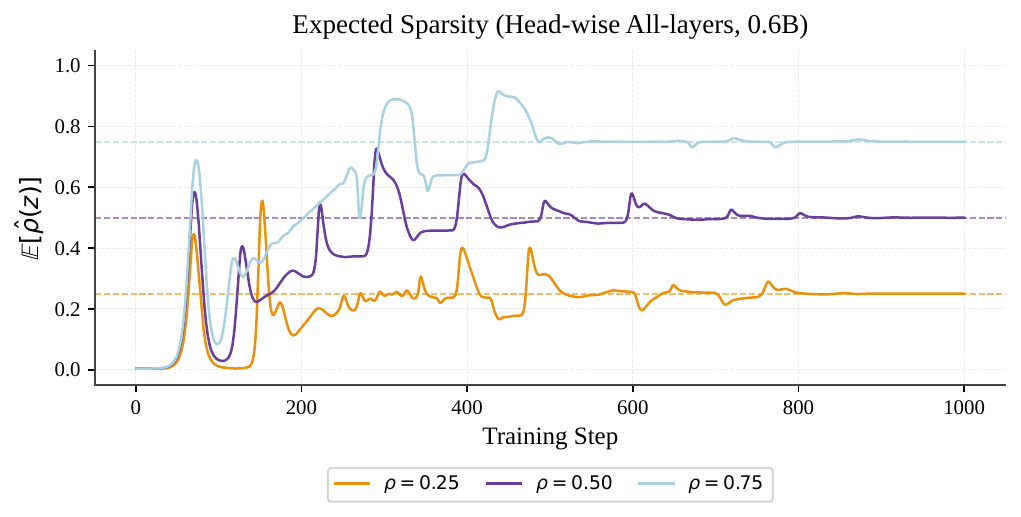}
        \caption{Head-wise (all-layers)}
        \label{fig:sparsity_hw_all}
    \end{subfigure}
    \caption{Trajectory of expected sparsity $\mathbb{E}[\hat{\rho}(z)]$ during Stage~1 mask learning on 0.6B at $\rho \in \{0.25, 0.50, 0.75\}$. Dashed horizontal lines indicate the target $\rho$. All configurations initially overshoot to a similar level before settling to their respective targets, with higher $\rho$ requiring less correction and thus converging earlier.}
    \label{fig:expected_sparsity}
\end{figure*}

\section{Additional Experimental Results and Analysis}

\subsection{Lagrangian Constraint Convergence}
\label{appdx:lagrangian-convergence}

We provide the full set of Lagrangian constraint loss trajectories that supplement the convergence analysis in Section~\ref{sec:convergence}. Figures~\ref{fig:lagrangian_06b_025}--\ref{fig:lagrangian_06b_075} present the constraint loss during Stage~1 mask learning on 0.6B at $\rho \in \{0.25, 0.50, 0.75\}$.

\paragraph{Granularity Affects Convergence Speed.}
The head-wise (single-layer) variant achieves the fastest convergence, with the constraint loss remaining close to zero from nearly the beginning of training.
This is because the per-layer constraint decomposes the global sparsity target into independent sub-problems, each involving only $H_{\mathrm{KV}}$ variables.
In contrast, the layer-wise variant and the head-wise (all-layers) variant both exhibit larger oscillations before stabilizing, with the latter showing the widest amplitude and the slowest convergence.
This phenomenon can be attributed to the enlarged search space of the global constraint, in which a single $\rho$ target is distributed across all $L \times H_{\mathrm{KV}}$ heads simultaneously.
The slower convergence of the head-wise (all-layers) variant is consistent with the lower downstream performance reported in Table~\ref{tab:main_result}.

\paragraph{Granularity Effects on 0.6B.}
At $\rho = 0.50$, the head-wise (single-layer) variant again converges the fastest among all three granularities, reproducing the pattern observed on 1.7B (Figure~\ref{fig:lagrangian_convergence}).
The 0.6B model exhibits a convergence profile highly similar to that of 1.7B, indicating that the optimization dynamics of the augmented Lagrangian are robust to model scale.

\paragraph{Convergence across Sparsity Levels on 0.6B.}
At all three sparsity levels, the constraint loss converges to near zero within 1,000 steps, confirming that \aname can precisely target arbitrary sparsity ratios.
The oscillation amplitude during the transient phase increases with $\rho$: the peak amplitude at $\rho = 0.75$ is roughly twice that at $\rho = 0.25$, reflecting the more aggressive redistribution required at higher sparsity.
Interestingly, convergence occurs earlier at higher $\rho$ despite the larger oscillations.
This behavior is apparent in the trajectory of the expected sparsity $\mathbb{E}[\hat{\rho}(z)]$ (Figure~\ref{fig:expected_sparsity}). 
During the early stages of training, the multipliers force $\mathbb{E}[\hat{\rho}(z)]$ to overshoot to a uniformly high level, irrespective of the target $\rho$. 
Consequently, a higher target necessitates less correction from this initial overshoot. 
In contrast, a lower target, such as $\rho = 0.25$, requires closing a larger gap to return to the designated value, thereby leading to slower convergence.

\subsection{Learned Allocation Patterns}
\label{appdendix:Patterns}
We provide the supplement to the analysis in Section~\ref{sec:pattern}.

\paragraph{FA Retreats Hierarchically as the SWA Budget Grows.}
A comparison of the three sparsity levels on 0.6B in Figure~\ref{fig:layerwise_025_075} shows how the model prioritizes the allocation of the FA budget.
At $\rho = 0.75$, where only 7 layers use FA, FA is restricted to two small clusters located near the early-middle and upper regions.
At $\rho = 0.50$, where 14 layers use FA, these clusters expand into larger blocks.
At $\rho = 0.25$, where 21 layers use FA, FA covers most of the model, but the bottom layers (0--2) and the final layer are still assigned to SWA.
This hierarchical retreat reveals a clear priority order: the middle-layer FA core is the last region to be replaced by SWA, indicating that it is the part of the network in which FA is most critical.

\paragraph{Intra-layer Heterogeneity under Head-wise Allocation.}
The head-wise heatmaps in Figure~\ref{fig:headwise_allocation} show that KV heads within the same layer often use different attention types.
At $\rho = 0.50$ on 0.6B, most layers contain a mixture of FA and SWA heads rather than being uniformly one type, confirming that layer-level decisions are suboptimal because they force a uniform attention type on heads that may serve functionally distinct roles.
Despite the finer granularity, the head-wise patterns preserve the same macro-level trend seen in layer-wise allocation: bottom layers are SWA-dominated, and middle layers are FA-dominated, as quantified by the per-layer SWA head ratio in Figure~\ref{fig:head_ratio}.

\subsection{Training FLOPs}
\begin{table}[t]
\centering
\small
\begin{tabular}{lcc}
\toprule
\textbf{Model} & \textbf{Dense FA} & \textbf{\aname{} ($\rho=0.50$)} \\
\midrule
0.6B & $17.42 \times 10^{15}$ & $14.65 \times 10^{15}$\,{\scriptsize \textcolor{customred}{($\downarrow$~15.9\%)}} \\
1.7B & $52.57 \times 10^{15}$ & $47.02 \times 10^{15}$\,{\scriptsize \textcolor{customred}{($\downarrow$~10.6\%)}} \\
\bottomrule
\end{tabular}
\caption{
Comparison of training FLOPs per step between Dense FA and \aname at $\rho=0.50$ under the Stage-2 setting.
The sequence length is $s = 8,192$, the global batch size is $512$, and the SWA window size is $w = 512$.
}
\label{tab:training_flops}
\end{table}

Table~\ref{tab:training_flops} reports the per-step training FLOPs of \aname and Dense FA on the 0.6B and 1.7B models under the Stage-2 configuration. 
At $\rho = 0.50$, \aname reduces per-step training FLOPs by $15.9\%$ on 0.6B and $10.6\%$ on 1.7B relative to Dense FA. 
Since attention FLOPs scale quadratically with sequence length, while the non-attention term scales linearly, the savings from \aname grow with context length, making the relative benefit more pronounced in long-context training regimes.

\subsection{Analysis of Attention Behavior}
\label{appendix:Attention Behavior}

\begin{figure*}[t]
    \centering
    \includegraphics[width=\textwidth]{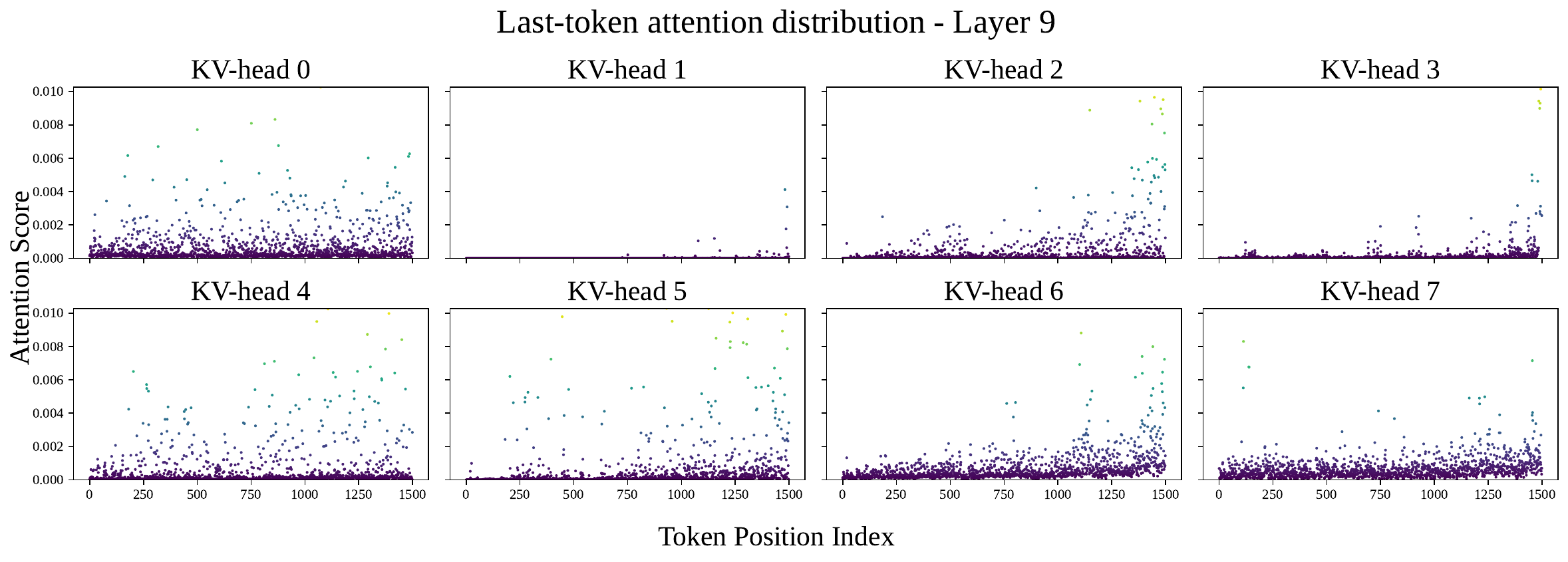}
    \caption{Head-wise last-token attention distribution for L9 of the
    0.6B model (assigned to FA across three $\rho$ in the layer-wise
    setting of Figure~\ref{fig:layerwise_025_075}).
    Most heads show \emph{dense broad} attention pattern.}
    \label{fig:headwise_L09}
\end{figure*}

\begin{figure*}[t]
    \centering
    \includegraphics[width=\textwidth]{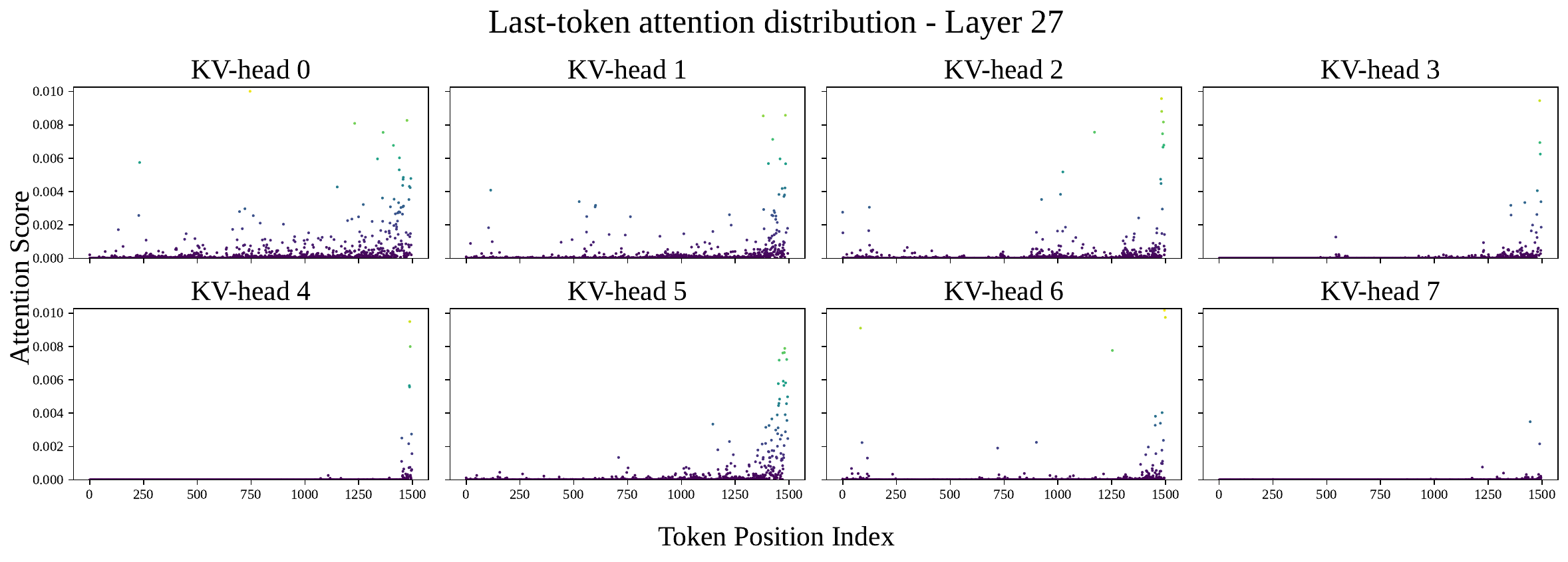}
    \caption{Head-wise last-token attention distribution for L27 of the
    0.6B model (assigned to SWA across three $\rho$ in the layer-wise
    setting of Figure~\ref{fig:layerwise_025_075}).
    Most heads show \emph{uniform} or \emph{local} attention pattern.}
    \label{fig:headwise_L27}
\end{figure*}

\paragraph{Task Selection and Evidence Distribution.}
We select samples from four subsets of LongBench~\cite{BaiLZL0HDLZHDTL24}, which together cover a range of evidence distributions.
NarrativeQA~\cite{KociskySBDHMG18} poses a question over a single long literary work, where the answer is typically supported by one or two localized passages.
HotpotQA~\cite{Yang0ZBCSM18} requires reasoning across multiple documents, so that the supporting evidence is spread over several non-contiguous regions of the context.
Passage Retrieval requires the model to identify which paragraph among many contains a given query.
In contrast to Needle-in-a-Haystack~\cite{LiuLHPBPL24}, in which the model must retrieve a short, semantically unrelated string from otherwise irrelevant context, the candidate paragraphs in Passage Retrieval are highly similar in topic.
This makes the task substantially more challenging and causes attention to be spread across multiple plausible paragraphs.
GovReport~\cite{HuangCPJW21} requires generating a summary of a government report, a task that depends on content distributed throughout the entire document.
For all visualizations, the input is truncated to a maximum length of 1{,}500 tokens.
The sample used in Figures~\ref{fig:casestudy_main}, \ref{fig:headwise_L09}, and~\ref{fig:headwise_L27} is drawn from NarrativeQA.
All attention behavior analyses presented are extracted from the pre-trained checkpoint before mask learning, and continued pre-training, so that the observed patterns reflect the model's natural attention behavior rather than adaptation to a particular FA/SWA configuration.

\begin{figure*}[t]
    \centering
    \begin{subfigure}[b]{0.48\textwidth}
        \centering
        \includegraphics[width=\textwidth]{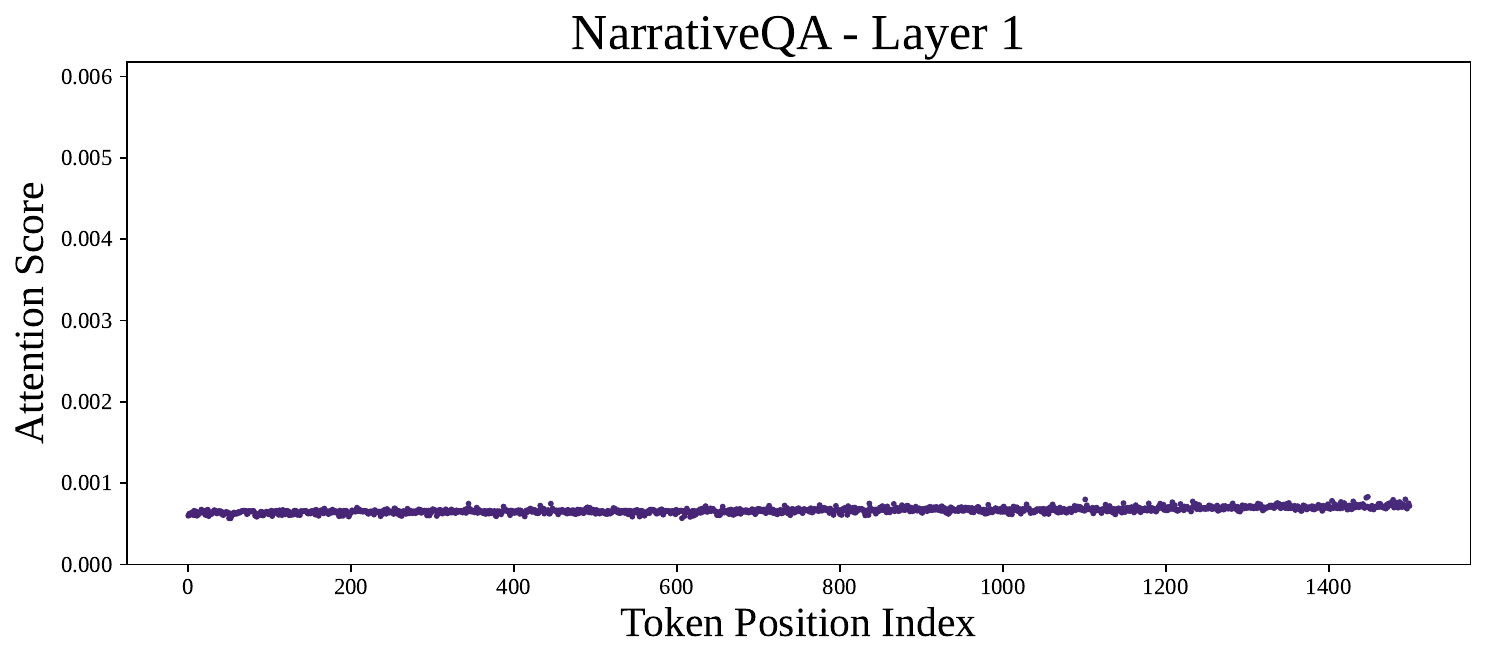}
    \end{subfigure}
    \hfill
    \begin{subfigure}[b]{0.48\textwidth}
        \centering
        \includegraphics[width=\textwidth]{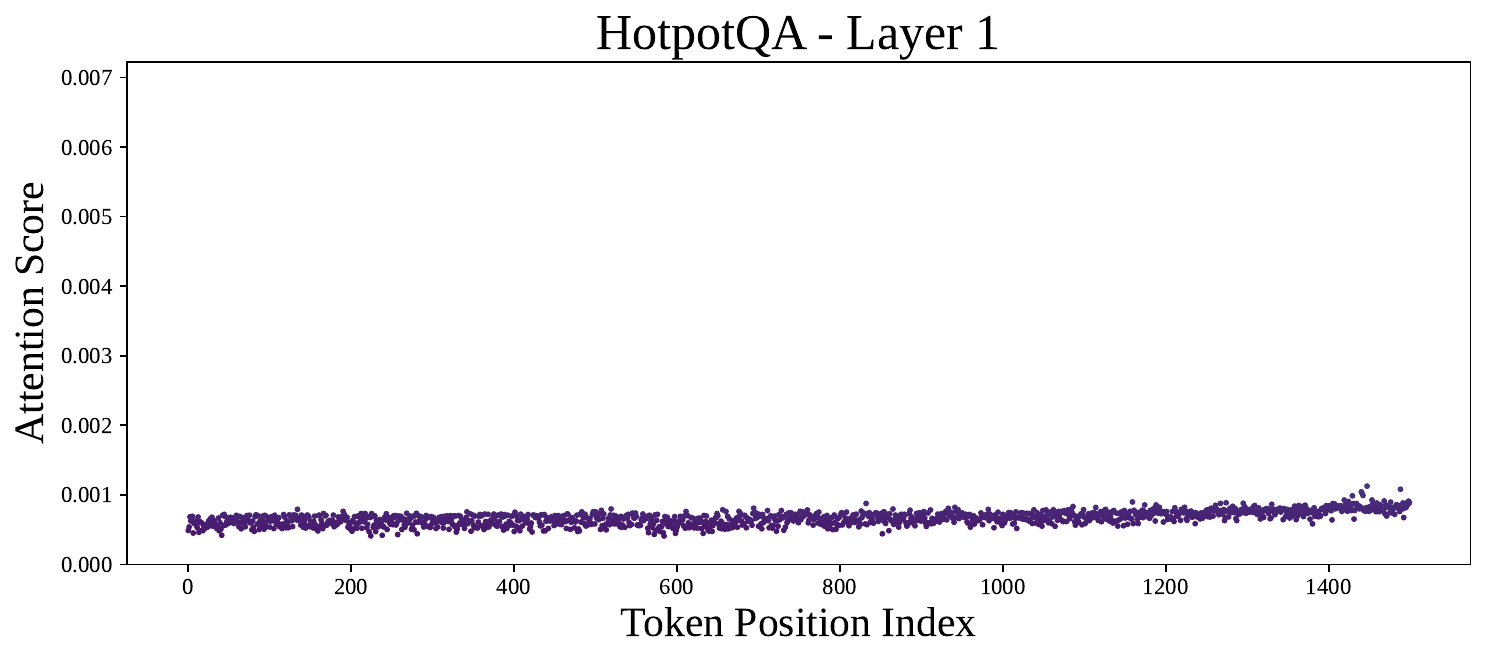}
    \end{subfigure}
    \vspace{0.3em}
    \begin{subfigure}[b]{0.48\textwidth}
        \centering
        \includegraphics[width=\textwidth]{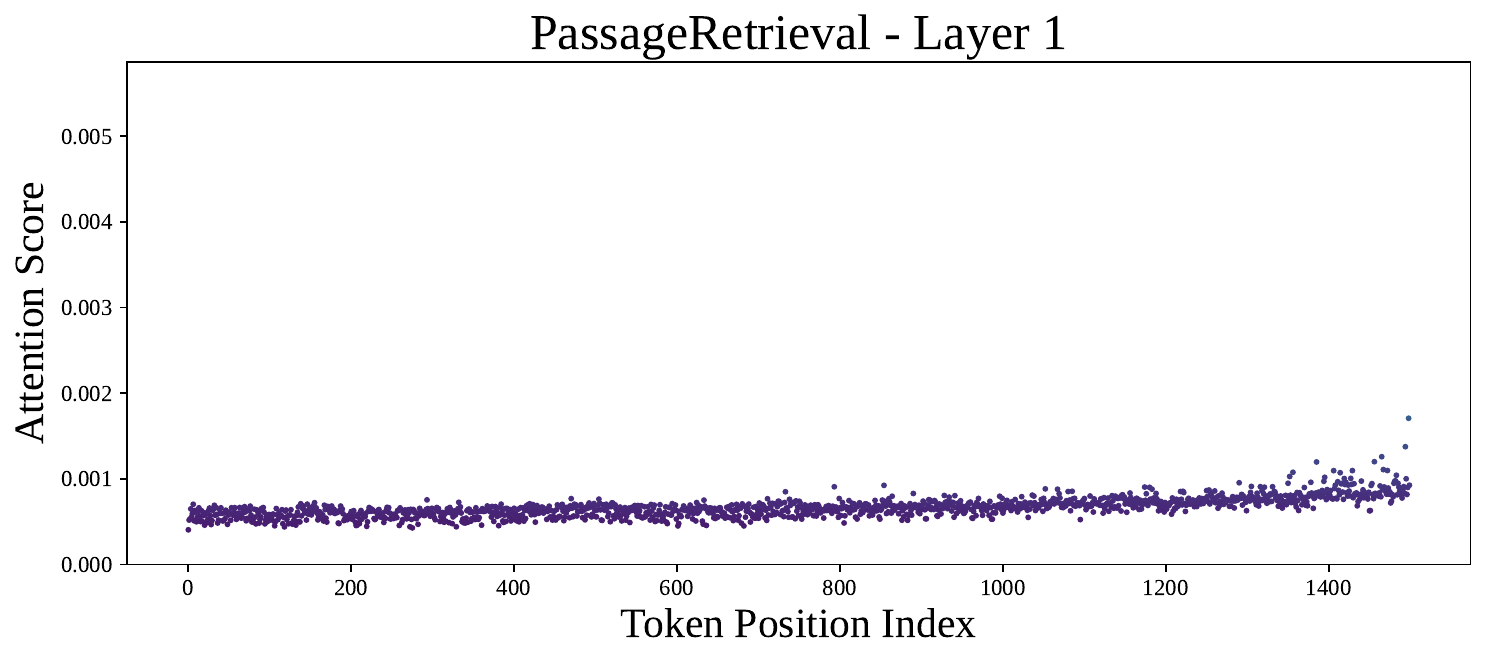}
    \end{subfigure}
    \hfill
    \begin{subfigure}[b]{0.48\textwidth}
        \centering
        \includegraphics[width=\textwidth]{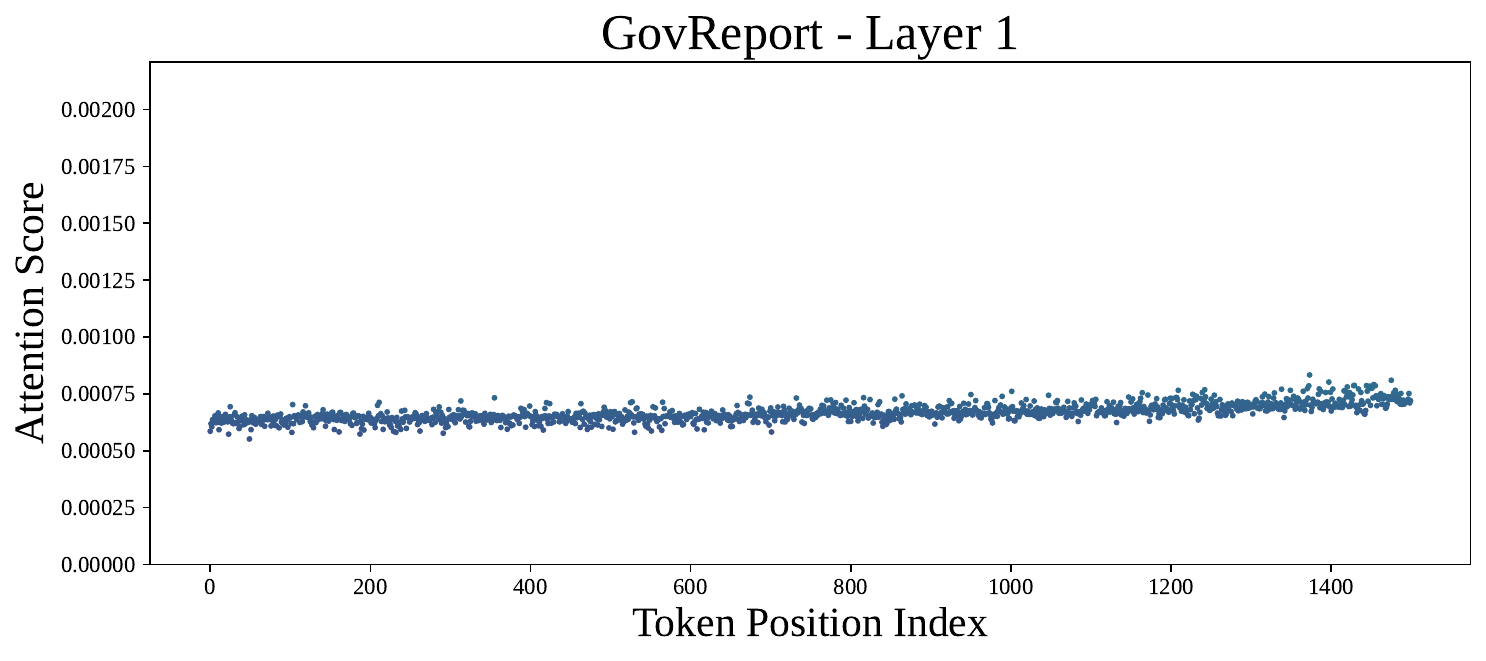}
    \end{subfigure}
    \caption{Cross-task last-token attention distribution for L1 of the 0.6B model.}
    \label{fig:crosstask_L01}
\end{figure*}

\begin{figure*}[t]
    \centering
    \begin{subfigure}[b]{0.48\textwidth}
        \centering
        \includegraphics[width=\textwidth]{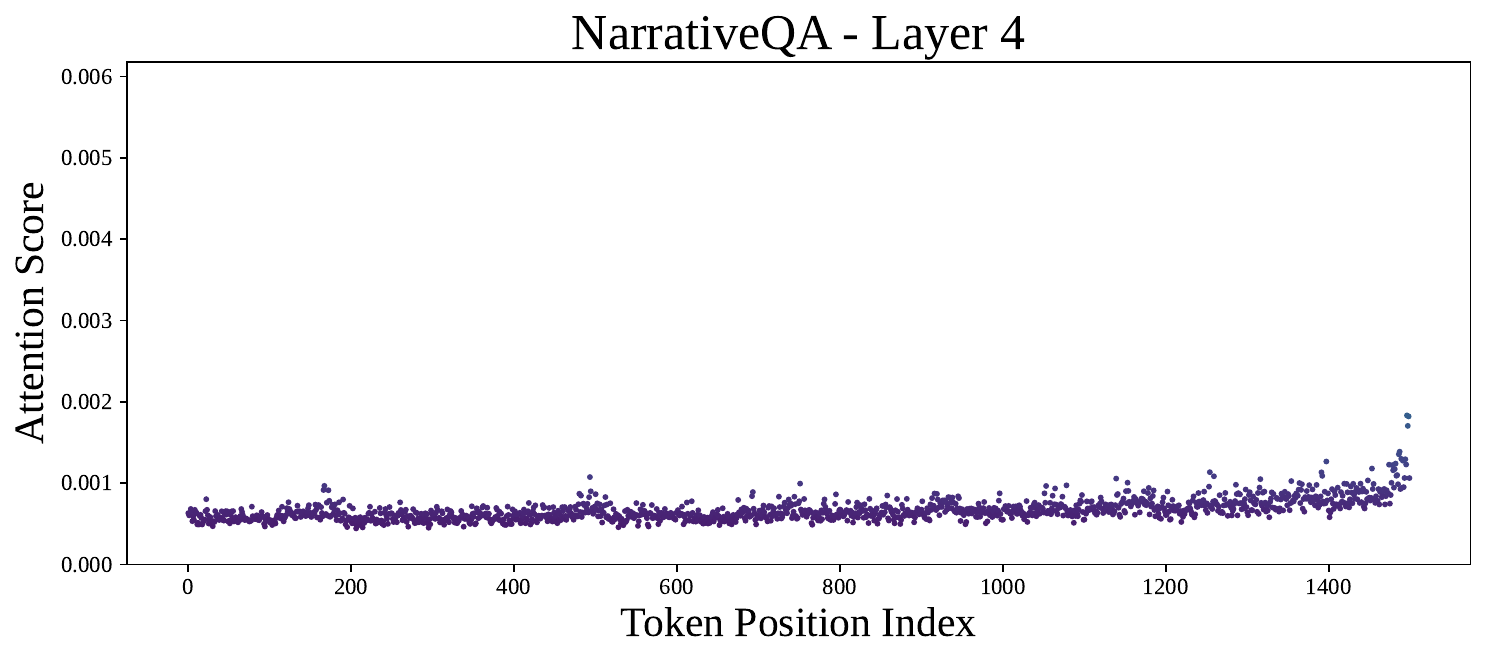}
    \end{subfigure}
    \hfill
    \begin{subfigure}[b]{0.48\textwidth}
        \centering
        \includegraphics[width=\textwidth]{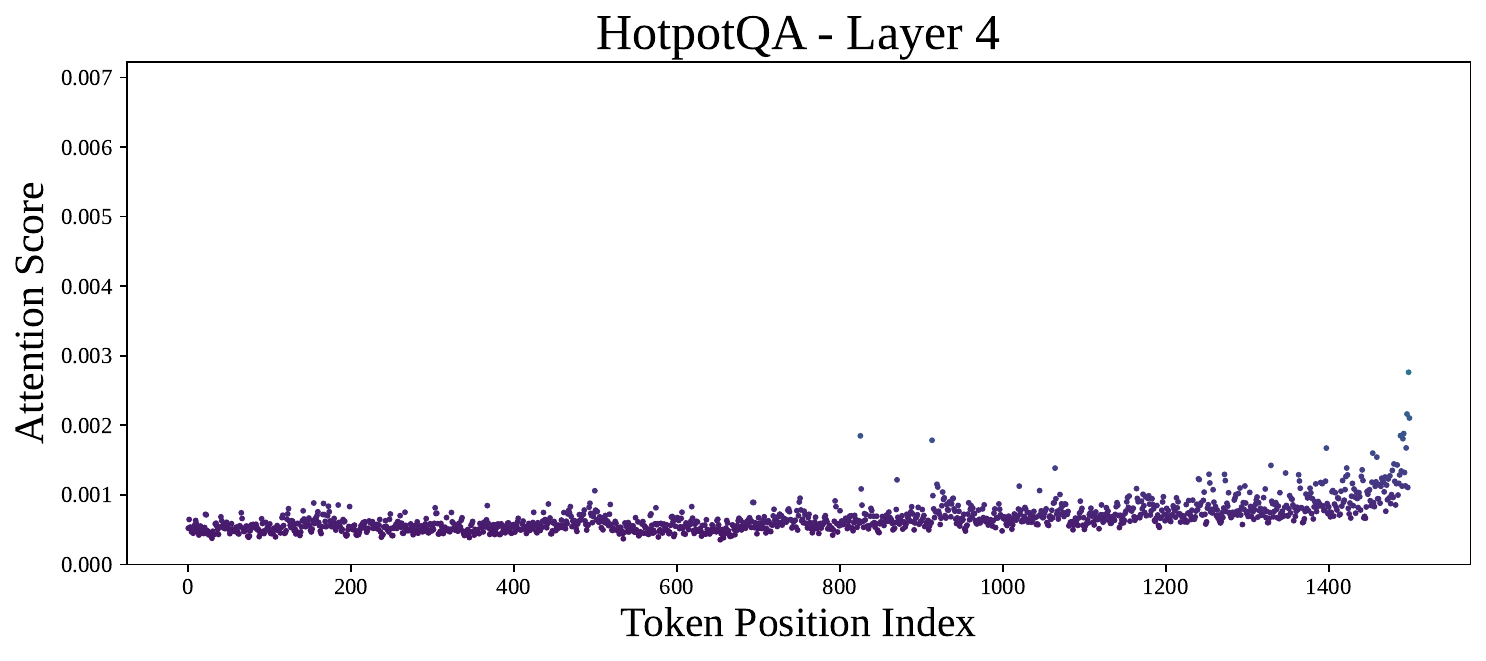}
    \end{subfigure}
    \vspace{0.3em}
    \begin{subfigure}[b]{0.48\textwidth}
        \centering
        \includegraphics[width=\textwidth]{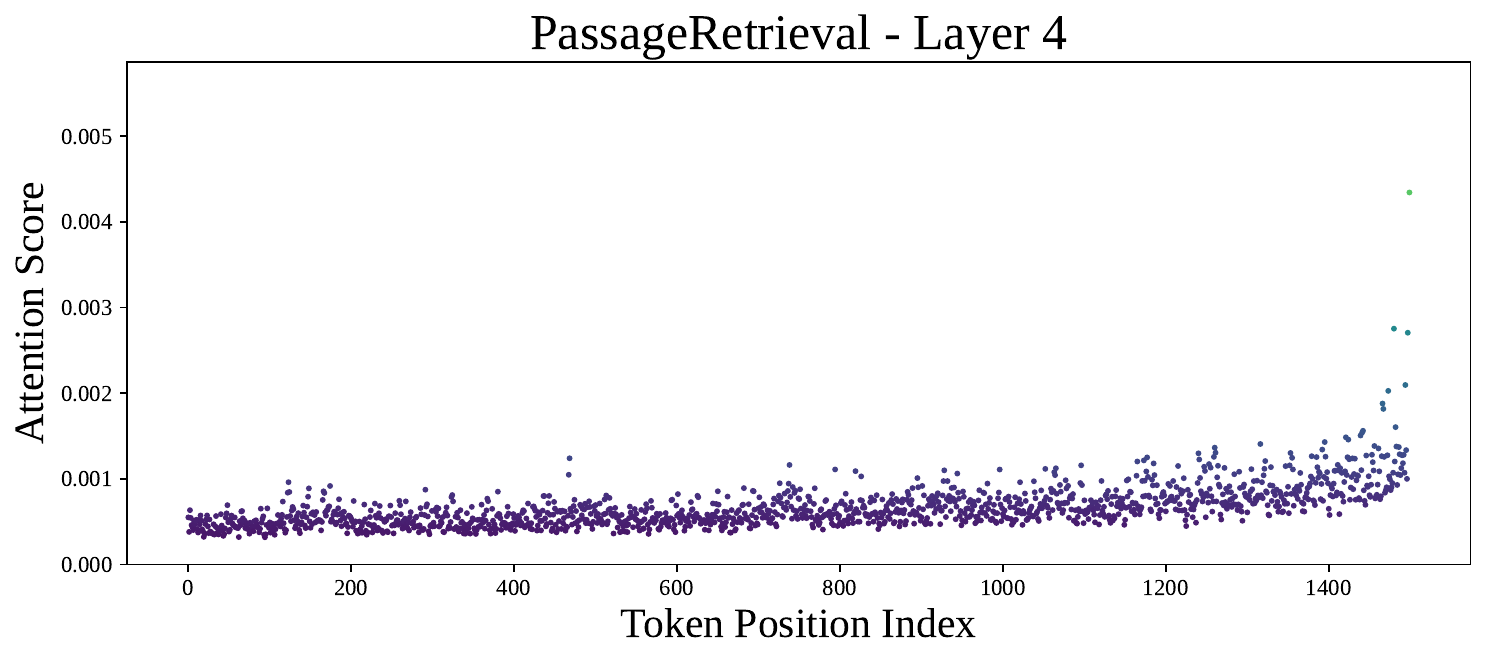}
    \end{subfigure}
    \hfill
    \begin{subfigure}[b]{0.48\textwidth}
        \centering
        \includegraphics[width=\textwidth]{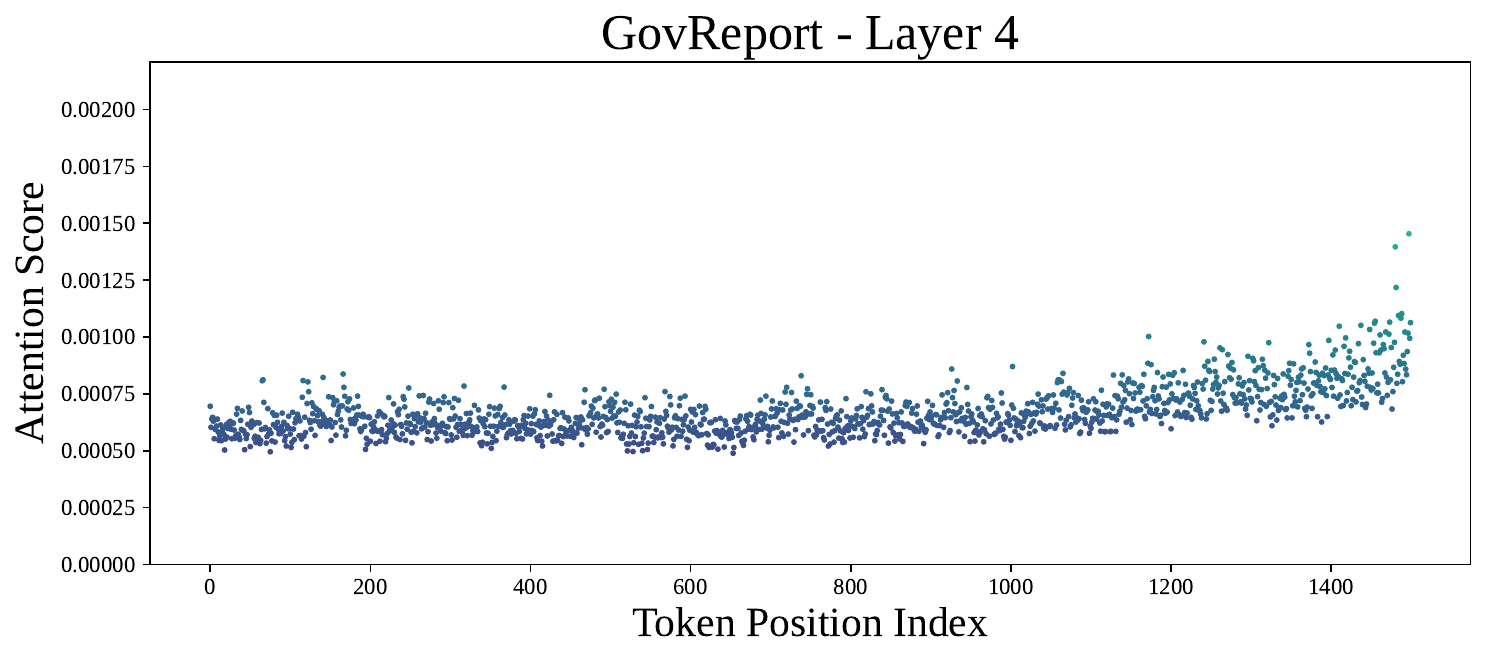}
    \end{subfigure}
    \caption{Cross-task last-token attention distribution for L4 of the 0.6B model.}
    \label{fig:crosstask_L04}
\end{figure*}

\begin{figure*}[t]
    \centering
    \begin{subfigure}[b]{0.48\textwidth}
        \centering
        \includegraphics[width=\textwidth]{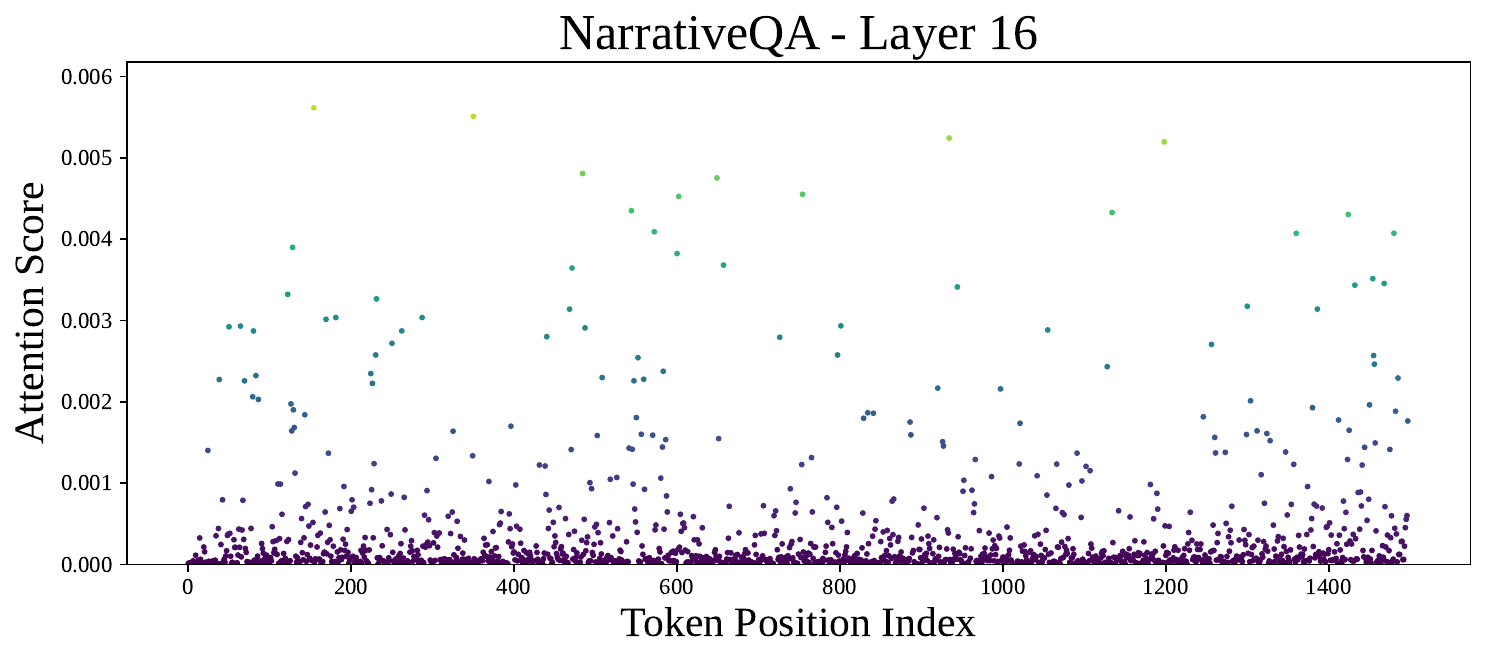}
    \end{subfigure}
    \hfill
    \begin{subfigure}[b]{0.48\textwidth}
        \centering
        \includegraphics[width=\textwidth]{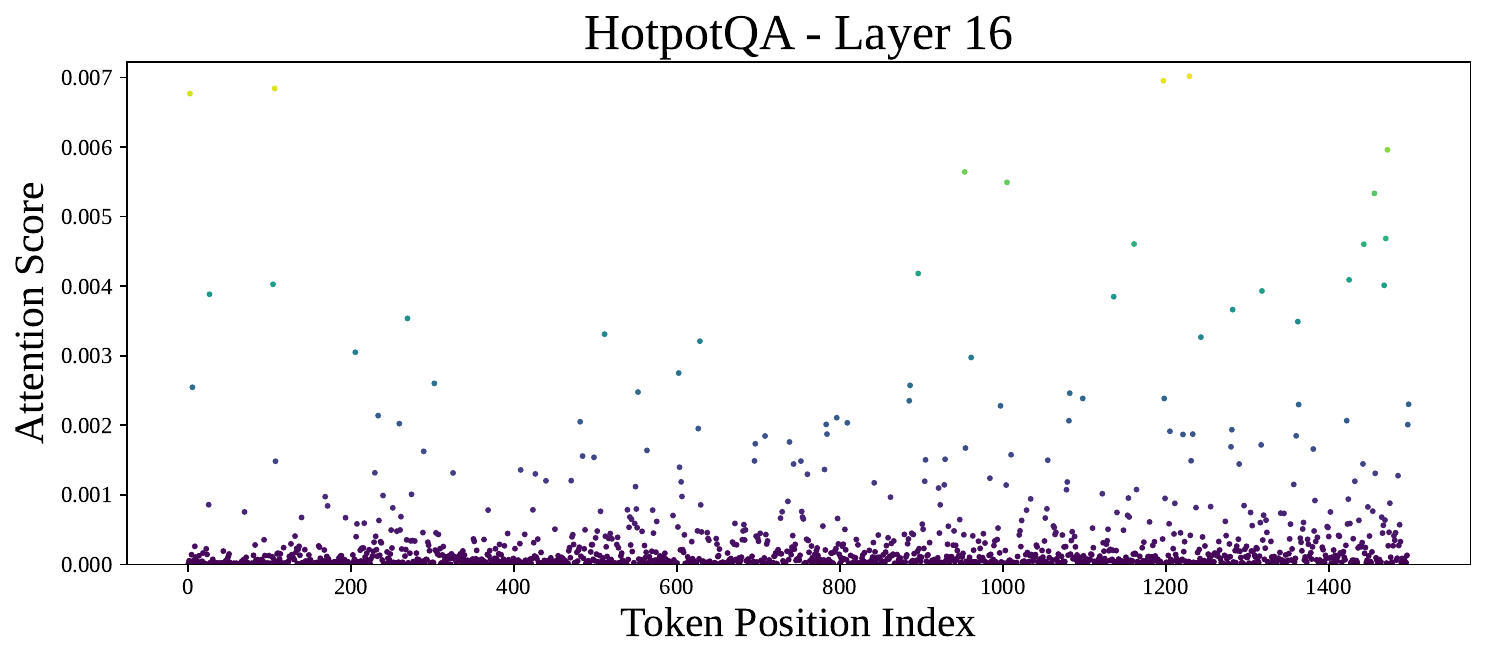}
    \end{subfigure}
    \vspace{0.3em}
    \begin{subfigure}[b]{0.48\textwidth}
        \centering
        \includegraphics[width=\textwidth]{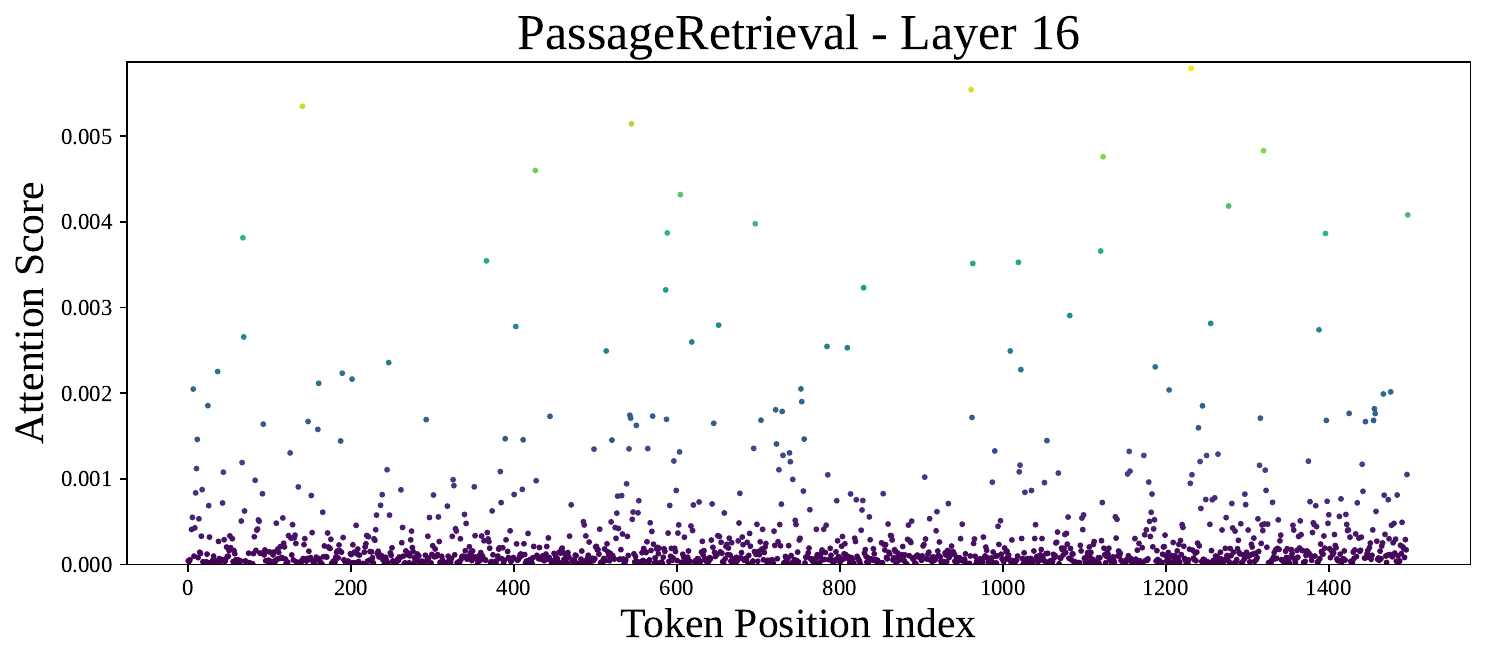}
    \end{subfigure}
    \hfill
    \begin{subfigure}[b]{0.48\textwidth}
        \centering
        \includegraphics[width=\textwidth]{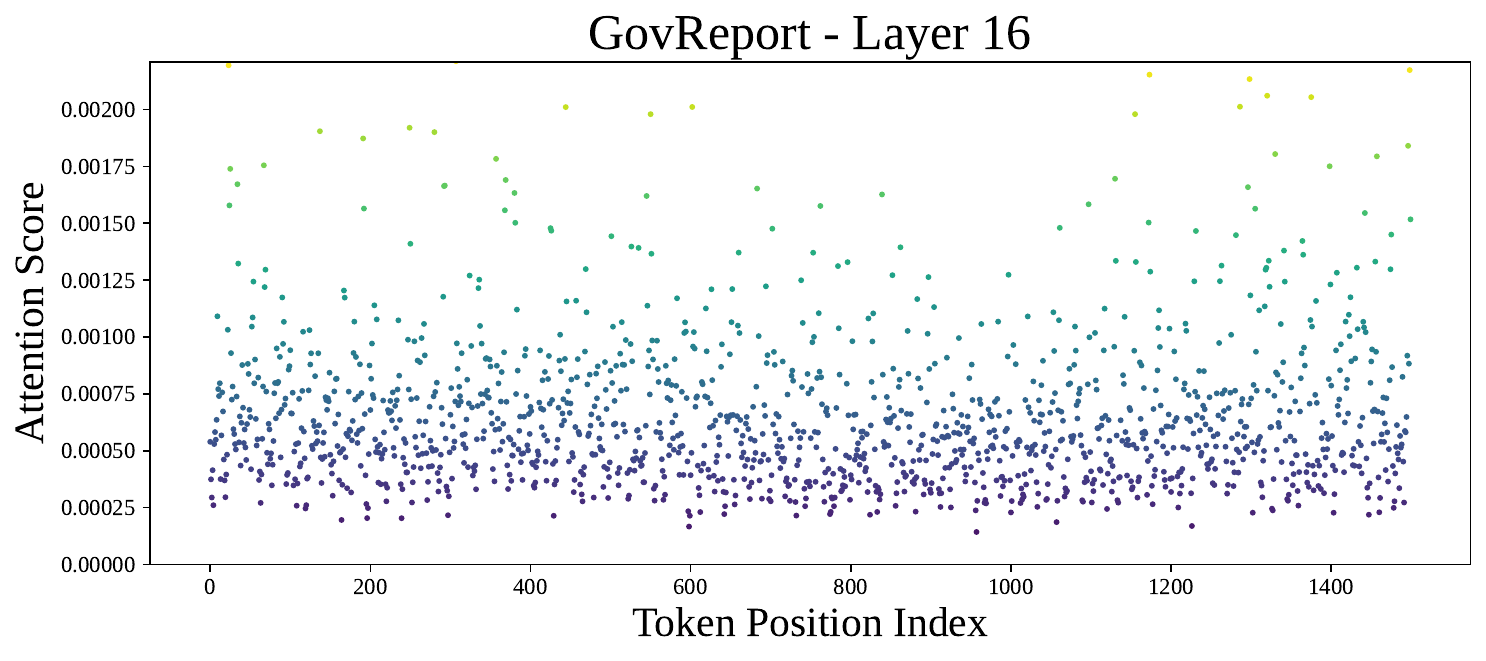}
    \end{subfigure}
    \caption{Cross-task last-token attention distribution for L16 of the 0.6B model.}
    \label{fig:crosstask_L16}
\end{figure*}

\begin{figure*}[t]
    \centering
    \begin{subfigure}[b]{0.48\textwidth}
        \centering
        \includegraphics[width=\textwidth]{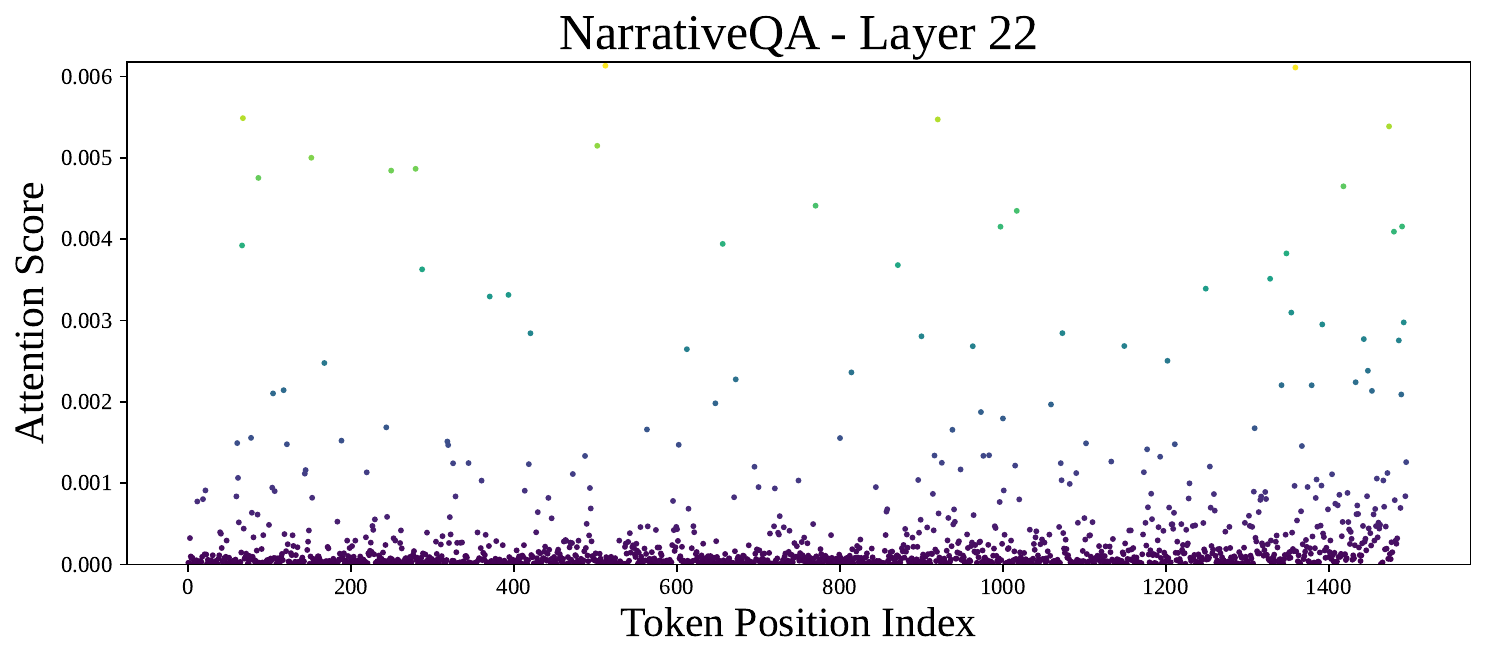}
    \end{subfigure}
    \hfill
    \begin{subfigure}[b]{0.48\textwidth}
        \centering
        \includegraphics[width=\textwidth]{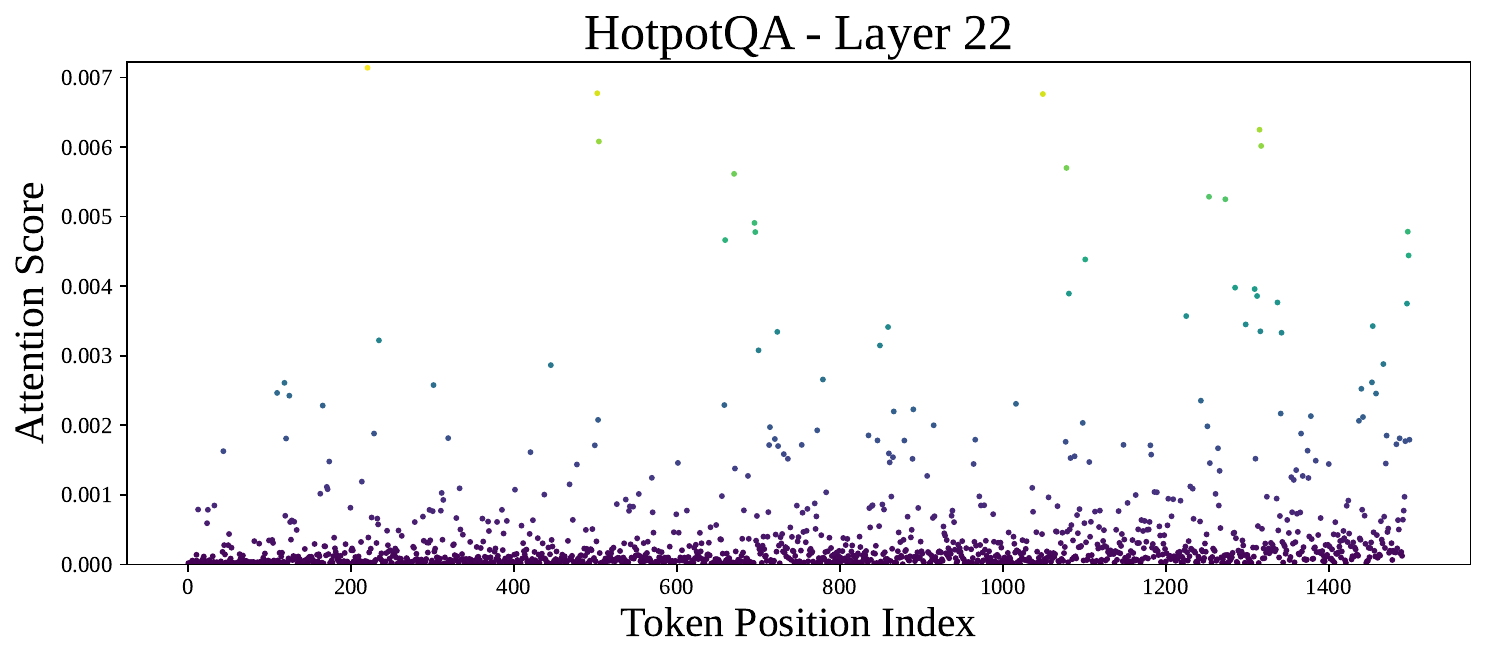}
    \end{subfigure}
    \vspace{0.3em}
    \begin{subfigure}[b]{0.48\textwidth}
        \centering
        \includegraphics[width=\textwidth]{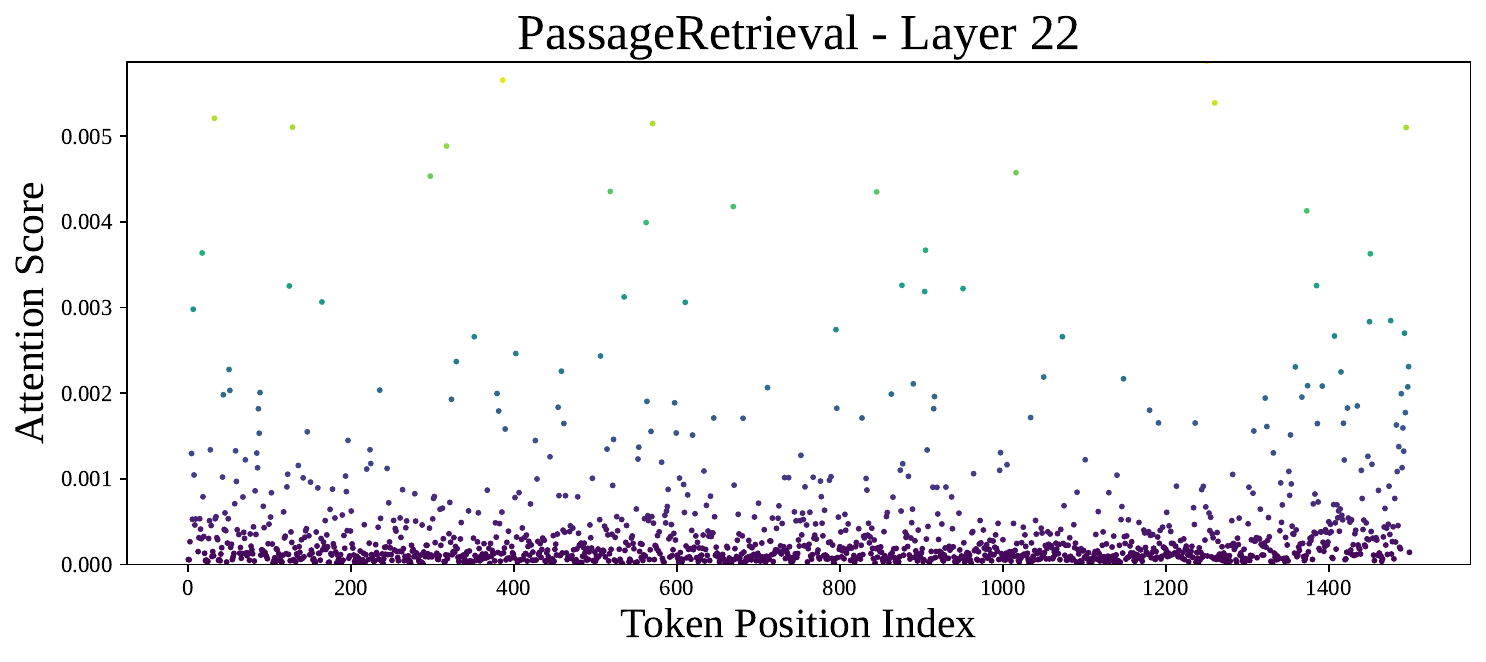}
    \end{subfigure}
    \hfill
    \begin{subfigure}[b]{0.48\textwidth}
        \centering
        \includegraphics[width=\textwidth]{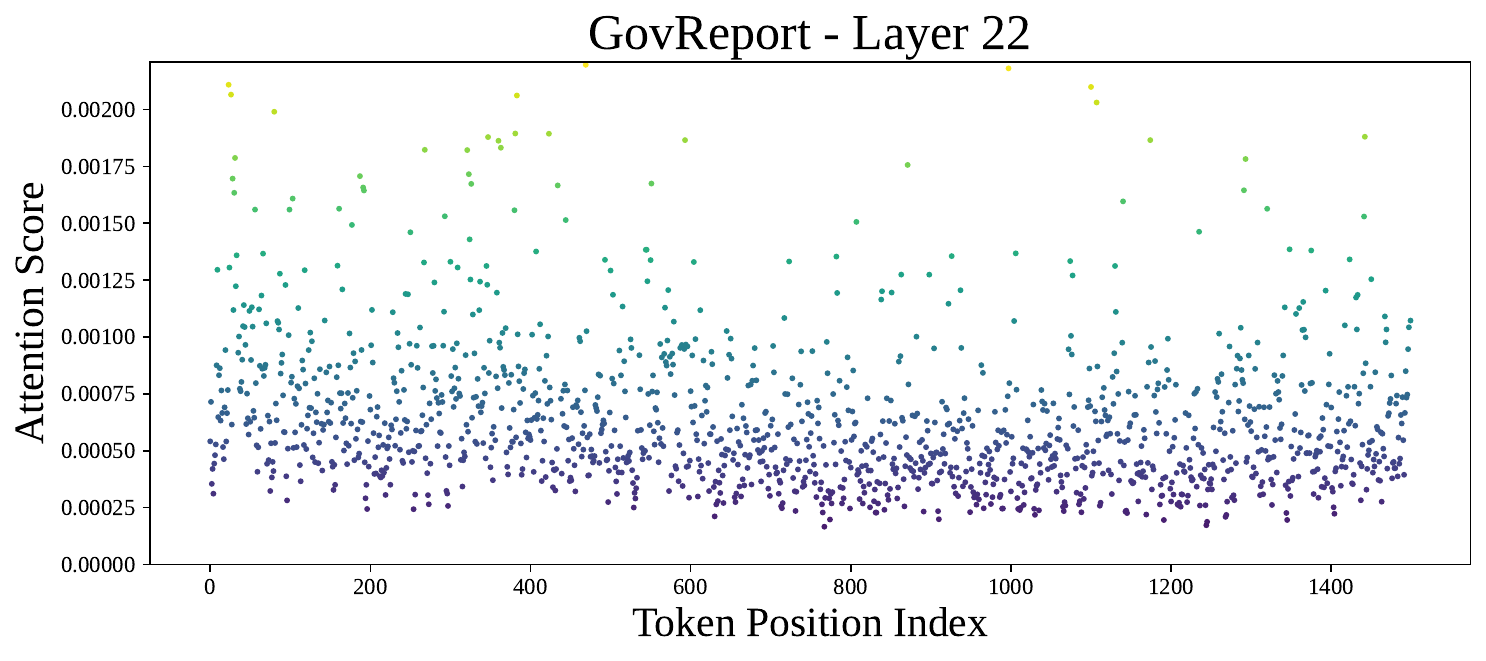}
    \end{subfigure}
    \caption{Cross-task last-token attention distribution for L22 of the 0.6B model.}
    \label{fig:crosstask_L22}
\end{figure*}

\paragraph{Cross-task Attention Behavior.}
We extend the analysis by examining how the same layers attend to inputs from different tasks (Figures~\ref{fig:crosstask_L01}--\ref{fig:crosstask_L22}).
The layers preferred for SWA across sparsity levels (L1, L4) exhibit only minor variation across tasks: L1 remains nearly uniformly distributed, and L4 retains its mild tail-side rise, with limited task-specific structure.
In contrast, the layers preferred for FA (L16, L22) show more pronounced cross-task differences: their spike distributions become denser and span a broader range as the evidence of the task becomes more dispersed, shifting from relatively concentrated spikes on question-answering inputs to broadly elevated attention across the full sequence on summarization inputs.
This suggests that the sparsity preferences of \aname effectively distinguish layers whose attention distribution is largely input-independent from those whose distribution adapts to the evidence structure of the task.
Preserving full attention is most valuable for the latter.

\begin{figure*}[t]
    \centering
    \begin{subfigure}[b]{0.45\textwidth}
        \centering
        \includegraphics[width=\textwidth,height=0.45\textheight,keepaspectratio]{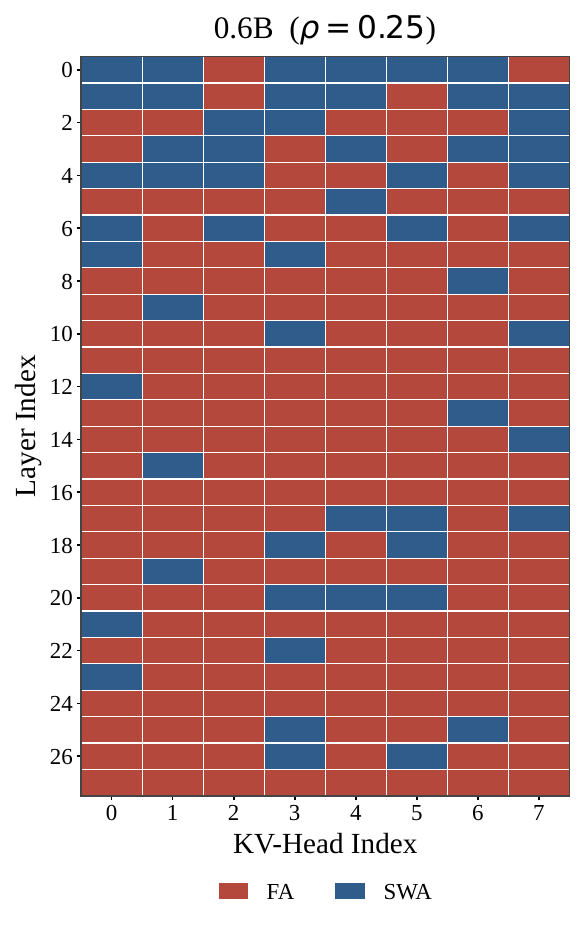}
        \caption{0.6B, $\rho = 0.25$}
        \label{fig:headwise_06B_025}
    \end{subfigure}
    \hfill
    \begin{subfigure}[b]{0.45\textwidth}
        \centering
        \includegraphics[width=\textwidth,height=0.45\textheight,keepaspectratio]{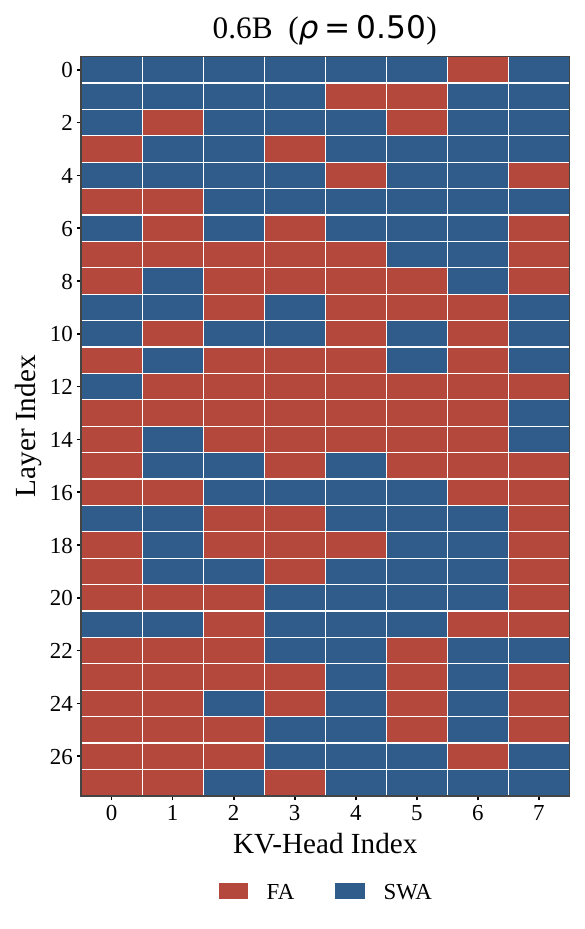}
        \caption{0.6B, $\rho = 0.50$}
        \label{fig:headwise_06B_050}
    \end{subfigure}

    \vspace{0.2em}

    \begin{subfigure}[b]{0.45\textwidth}
        \centering
        \includegraphics[width=\textwidth,height=0.45\textheight,keepaspectratio]{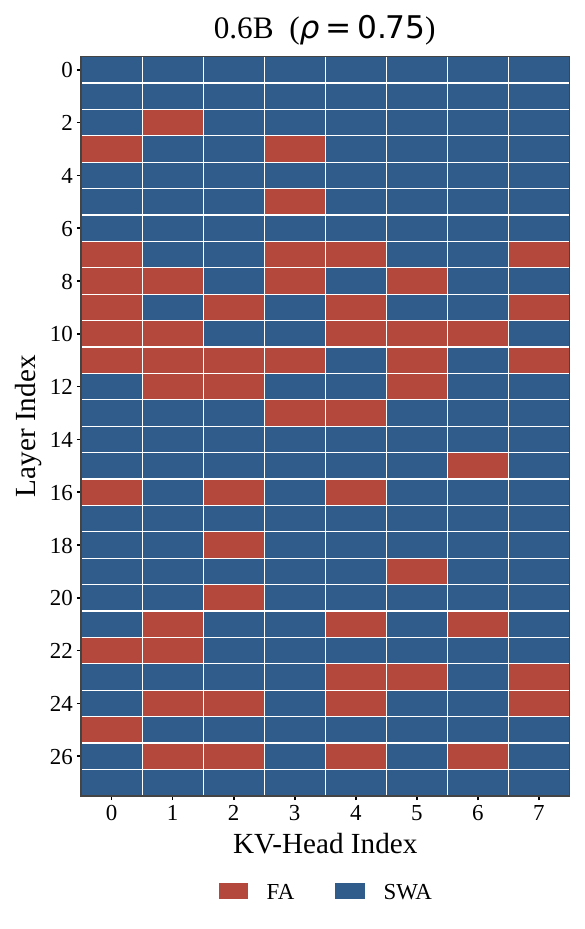}
        \caption{0.6B, $\rho = 0.75$}
        \label{fig:headwise_06B_075}
    \end{subfigure}
    \hfill
    \begin{subfigure}[b]{0.45\textwidth}
        \centering
        \includegraphics[width=\textwidth,height=0.45\textheight,keepaspectratio]{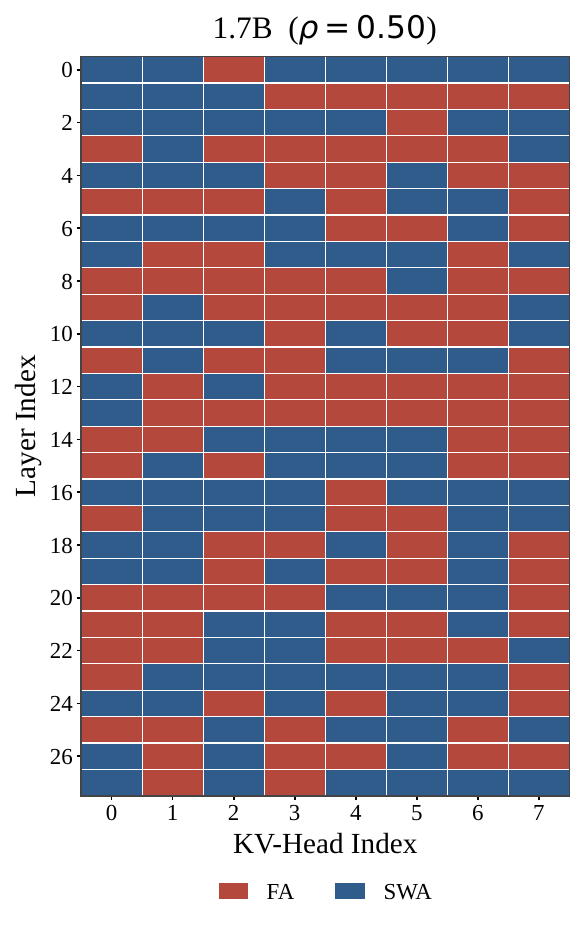}
        \caption{1.7B, $\rho = 0.50$}
        \label{fig:headwise_17B}
    \end{subfigure}

    \caption{
    Learned head-wise FA/SWA allocation across model scales and sparsity levels.
    Each cell indicates whether a KV head uses \textcolor{customred}{FA (red)} or \textcolor{customblue}{SWA (blue)}.
    }
    \label{fig:headwise_allocation}
\end{figure*}

\paragraph{Intra-layer Head Heterogeneity.}
The layer-wise visualization averages across all KV heads within a
layer, potentially masking divergent behaviors among individual heads.
Figures~\ref{fig:headwise_L09} and~\ref{fig:headwise_L27} decompose two representative layers into their eight KV heads.
In L9, which is assigned to FA across three sparsity levels in the
layer-wise setting, the majority of heads exhibit \emph{dense broad} attention, but others remain nearly uniform.
The layer-wise decision is dominated by the broad-attending majority,
but the uniform heads gain little from full attention.
L27 presents the mirror case: assigned to SWA across all sparsity
levels in the layer-wise setting, most of its heads are uniform or
local, but a minority display distant spikes that the SWA window would truncate.
Here, the layer-wise decision is dominated by the local majority, at the
cost of discarding the long-range information captured by the few broad heads.
Accordingly, under head-wise allocation
(Figure~\ref{fig:headwise_allocation}), the heads within L9 and L27 are
no longer forced into a single type; instead, as $\rho$ increases,
individual heads progressively transition to SWA based on their own
attention range, in contrast to the layer-wise setting
(Figure~\ref{fig:layerwise_025_075}) where the entire layer switches
at once.
This heterogeneity explains why head-wise allocation outperforms
layer-wise allocation in Table~\ref{tab:main_result}: head-wise
granularity allows \aname to retain FA selectively for the broad heads
within an otherwise local layer and, conversely, to release uniform
heads within an otherwise broad layer.

% case study

\end{document}